\documentclass[10pt,twocolumn,letterpaper]{article}

\usepackage{wacv}              % To produce the CAMERA-READY version
\usepackage[accsupp]{axessibility}  % Improves PDF readability for those with disabilities.

\usepackage{graphicx}
\usepackage{amsmath}
\usepackage{amssymb}
\usepackage{booktabs}
\usepackage{makecell}
\usepackage[table]{xcolor}
\usepackage{color}
\usepackage{array,multirow}
\usepackage{bbm}

\usepackage[pagebackref,breaklinks,colorlinks]{hyperref}

\usepackage[capitalize]{cleveref}
\crefname{section}{Sec.}{Secs.}
\Crefname{section}{Section}{Sections}
\Crefname{table}{Table}{Tables}
\crefname{table}{Tab.}{Tabs.}

\DeclareMathOperator*{\argmax}{arg\, max}  

\definecolor{aliceblue}{rgb}{0.94, 0.97, 1.0}

\def\wacvPaperID{1539} % *** Enter the WACV Paper ID here
\def\confName{WACV}
\def\confYear{2024}

\begin{document}

\title{Unsupervised Domain Adaptation for Semantic Segmentation\\ with Pseudo Label Self-Refinement}

\author{{Xingchen Zhao$^{2*}$,~ Niluthpol Chowdhury Mithun$^{1}$}\thanks{\textsuperscript{}Equal Contribution},~ Abhinav Rajvanshi$^{1}$, \\ Han-Pang Chiu$^{1}$, ~Supun Samarasekera$^{1}$ \and
{\normalsize $^{1}$SRI International, Princeton, NJ, USA} \and {\normalsize $^{2}$Northeastern University, Boston, MA, USA} 
\and
\texttt{\small $^{1}$firstname.lastname@sri.com} \and \texttt{\small $^{2}$zhao.xingc@northeastern.edu}
}
\maketitle

%%%%%%%%% ABSTRACT
\begin{abstract}
   Deep learning-based solutions for semantic segmentation suffer from significant performance degradation when tested on data with different characteristics than what was used during the training. Adapting the models using annotated data from the new domain is not always practical. Unsupervised Domain Adaptation (UDA) approaches are crucial in deploying these models in the actual operating conditions. Recent state-of-the-art (SOTA) UDA methods employ a teacher-student self-training approach, where a teacher model is used to generate pseudo-labels for the new data which in turn guide the training process of the student model. Though this approach has seen a lot of success, it suffers from the issue of noisy pseudo-labels being propagated in the training process. To address this issue, we propose an auxiliary pseudo-label refinement network (PRN) for online refining of the pseudo labels and also localizing the pixels whose predicted labels are likely to be noisy. Being able to improve the quality of pseudo labels and select highly reliable ones, PRN helps self-training of segmentation models to be robust against pseudo label noise propagation during different stages of adaptation. We evaluate our approach on benchmark datasets with three different domain shifts, and our approach consistently performs significantly better than the previous state-of-the-art methods.
\end{abstract}

%%%%%%%%% BODY TEXT
\section{Introduction}

Semantic segmentation, a well-studied computer vision task, has seen significant advances with deep neural networks in the last decade~\cite{chen2017deeplab,xie2021segformer,pspnet,robinSegmenter,adaptive_pyramid, tian2022striking}. In practice, these models rely strongly on large-scale annotated datasets for training. However, the characteristics of the datasets used for training could be significantly different from those in the actual operational scenarios, e.g., changes in camera sensors, and lighting conditions. When there is a distribution shift between train (i.e., source) and test (i.e., target) sets, model accuracy often degrades dramatically ~\cite{csurka2017comprehensive,patel2015visual,wang2018deep,csurka2021unsupervised}. Creating new annotated datasets for retraining is costly, especially for per-pixel annotation.

To alleviate this issue, various UDA approaches have been developed over recent years, which focus on adapting models trained from a source domain to target domains with unlabeled data~\cite{csurka2021unsupervised,hoyer2022daformer,zhang2021prototypical,karim2023c}. For example, many benchmark manually annotated outdoor driving semantic segmentation datasets (e.g., Cityscapes~\cite{cordts2016cityscapes} daytime driving dataset, SYNTHIA~\cite{ros2016synthia} synthetic driving dataset) are available to train high-performing neural network models for the source domain, but only unlabeled data is available for the target domain (e.g., nighttime driving dataset Dark Zurich~\cite{sakaridis2019guided}). State-of-the-art UDA semantic segmentation techniques often use a teacher-student self-training approach, iteratively training a student model with pseudo-labeled target data generated by a teacher model~\cite{tranheden2021dacs,hoyer2022daformer,bruggemann2023refign}. While self-training approaches have demonstrated their effectiveness, they suffer significantly from the erroneous model prediction propagation issue, i.e., confirmation bias~\cite{tarvainen2017mean,yang2022divide}. Pseudo-labels are very likely to be noisy especially during the early stages of training due to the source-target domain gap. If the issue is not addressed, it consequently leads to corrupted models with degraded generalization performance. 

We propose to train an auxiliary neural network model for refining the pseudo labels. Our proposed pseudo-label refinement network (PRN) is trained to serve two main purposes: it refines noisy pseudo labels, improving their quality, and localizes potential errors in pseudo labels by predicting a binary mask for challenging pixels (that are likely to have incorrectly predicted labels). The first task focuses on correcting pseudo-labels, while the second task helps in pseudo-label selection. 
%perform two complementary tasks. The first is to rectify the noisy pseudo labels, i.e., produce refined higher-quality labels. The second is to localize possible errors in pseudo-label prediction, i.e., predict a binary mask indicating difficult pixels in the image which are likely to have incorrectly predicted labels. Here, the second task mainly focuses on pseudo-label selection, whereas the first task focuses on correcting pseudo-labels. 
Note that it is possible to just carry out the first task and to select pixels with maximum softmax probability (of corrected segmentation logits) below a selected threshold as the erroneous pseudo labels. However, the performance of such a naive approach would be very sensitive to the selected threshold. Moreover, it is evident that the first task is class-specific, whereas the second is class-agnostic. Therefore, we train our model specifically for the task of pseudo-label error mask prediction, which also helps to learn effective representations to correct erroneous pseudo labels. PRN minimizes confirmation bias, making self-training for semantic segmentation models more robust against noisy pseudo-labels. 
%PRN is trained by taking in inputs of  perturbed segmentation labels (i.e., logits) as well as features for a sample and  learning to predict refined segmentation labels for the sample. The pseudo-label refinement network can be trained only using the labeled source data via supervised learning. However, such a model trained with only source domain images is unlikely to learn to effectively refine the target domain image pseudo labels. 
%Moreover, effective training of PRN also depends on perturbation quality and we find PRN model trained with pseudo-label perturbed with random noise is ineffective in improving performance over the self-training baseline.  In this regard, we develop a Fast Fourier Transform (FFT) based perturbation strategy that creates perturbed segmentation labels for a sample from one domain replacing its low-frequency part of the amplitude with the amplitude of a random sample from the other domain and keeping the phase unaltered. As phase encodes high-level semantics of a sample~\cite{yang2020fda} which is not changed, it allows using ground-truth (GT) label of the sample for training. However, due to the injection of the low-frequency part of amplitude, the perturbed sample now contains some style characteristics of the other domain. The same process is adopted to perturb both source and target pseudo-labels.  We hypothesize, by refining pseudo-labels perturbed using this process, PRN model will learn generalizable features across domains and ultimately be able to effectively refign target pseudo-labels.

Our PRN model takes the noisy segmentation logits from the teacher decoder and image features from the teacher encoder as the input and predicts the refined logits and the noise mask. This ability of our model is achieved by employing a novel training strategy using Fast Fourier Transform (FFT) based perturbations. Fourier transformation can be applied to decompose any signal (e.g., RGB image, features, logits) to amplitude (i.e., intensities or style) and phase (i.e., spatial positions or semantics) components \cite{piotrowski1982demonstration}.
%It has been observed in prior studies~\cite{yang2020fda,zhao2022test} that the low-level spectrum (amplitude) encodes the style characteristics whereas the phase encodes the high-level semantics. 
We perturb the amplitude of source image segmentation logits using the amplitude of a random target image to effectively introduce noise while preserving object structure, facilitating effective learning of the PRN model. It allows the use of ground truth (GT) labels for the source data as supervision for the model, while style information from the target domain also acts as training inputs through perturbed logits.
%As perturbed logits contain the style from the target domain, they act as training inputs for our network. Since ground truth (GT) labels for the source data are available, they can be used to provide supervision. 
%Note that we can only use labeled source data to train our network. 
%However, such a model trained with only source domain images is unlikely to learn to effectively refine noisy target logits. 
We also train our model using target domain data (perturbed with source style) with a similar process. Since access to target GT labels is not available, pseudo-labels of unperturbed target logits are used as supervision. This training strategy helps the model learn robust features across domains to effectively refine target pseudo-labels.

\textbf{Contributions: }Our work aims to tackle the propagation of noisy pseudo-labels in the training process. It has two main contributions. The first is a new pseudo-label refinement module that learns to predict the refined pseudo labels as well as the error mask containing the information about noisy labels. Our approach is different from previous approaches that rely on threshold-based selective pseudo-labeling. We also developed a novel training strategy using FFT-based perturbations that enables us to achieve the desired behavior of the refinement module. As the second contribution, our framework outperforms SOTA methods significantly in three UDA segmentation benchmarks, covering normal-to-adverse weather and synthetic-to-real adaptation. 

%-------------------------------------------------------------------------
\section{Related Works}

%\subsection{Semantic Segmentation}
%Semantic segmentation is the task of assigning a semantic class to every pixel of a given image. Various approaches have been developed over the last decade for performing this task effectively. Convolutional Neural Networks (CNN) based models have been the most popular choice for this task over the last decade.
%as these models can capture the context around a pixel in an image. However, the context in these models is limited to the neighboring regions around a pixel. There have been numerous efforts for improving the performance of these models by integrating more contextual information within the models. These efforts include Graphical models \cite{marvinCRF, sidCRF}, pyramid network based models \cite{tsung_pyramid, lap_pyramid, pspnet, adaptive_pyramid}, attention-based models \cite{chen_attention, pyramid_attention, dual_attention}, and dilated convolutional models \cite{chen2017deeplab, liang_rethinking, deeplabv3}. 
%All of these architectures attempted to boost the performance of CNNs, by capturing more contextual information in different ways. Recently, Transformer-based models \cite{guoSOTR, robinSegmenter, xie2021segformer} have gained attention in the field of computer vision. These models are capable of capturing the global context throughout the network, and hence have achieved state-of-the-art performance in image segmentation. We also choose a transformer-based backbone for our model due to its superior performance.

\subsection{Unsupervised Domain Adaptation}

%The problem of adapting a deep neural network model to a new domain has been around for a long time and has been widely studied in the literature. 
A number of unsupervised domain adaptation techniques~\cite{udaSurvey, csurka2017comprehensive} have been developed for reducing the domain gap between the source and target data, specifically for the semantic segmentation task. For example, Distribution Discrepancy Minimization \cite{karstenMMD} seeks to minimize the distribution discrepancy between source and target domains in some latent feature space. Curriculum learning \cite{yangCurriculum, sakaridis2019guided, qingCurriculum} has also been used for domain adaptation which involves learning easier tasks before more complex tasks. Self-ensembling \cite{choiSelfEns, wangSelfEns, chrisSelfEns} uses an ensemble of models, and exploits the consistency between predictions under some perturbations. Adversarial training \cite{jihanAdv, jinyuAdv, jiaxingAdv} is another popular approach that achieves the same goal by training with both clean and adversarial samples. Recently, self-training has been the most popular method for UDA  \cite{zou2018unsupervised, hoyer2022hrda, hoyer2022daformer, tranheden2021dacs, li2019bidirectional, myeongTexture, guanContent, zhongDiff, bruggemann2023refign}, in which the pseudo-labels are generated for the target-data (typically by a teacher model) and then used to train the target domain model. This approach has shown SOTA performance for the UDA semantic segmentation task. However, it generally suffers from the presence of significant noise in the pseudo-labels. 

\subsection{Pseudo Label Refinement}

The potential existence of noisy pseudo-labels in the self-training method is likely to result in subpar performance~\cite{arpit2017closer}. Hence, the key concept for these methods revolves around producing dependable pseudo-labels. Some works in semi-supervised learning address this by employing a neural network module to rectify pseudo-labels or identify errors~\cite{kwon2022semi,ke2020guided,mendel2020semi}. In their context, both labeled and unlabeled data stem from the same domain, allowing the refinement module to be trained using labeled data. However, it is not applicable to UDA, where labeled and unlabeled data pertain to distinct domains. Among prior UDA semantic segmentation works, most employ selective pseudo-labeling (e.g.,\cite{wang2021uncertainty,subhani2020learning}). Some of the works \cite{guanContent, zhongDiff, zou2018unsupervised} rely on the softmax of the model output as a confidence measure. \cite{mei2020instance} uses an adaptive confidence threshold that is updated throughout the training, while \cite{zhengRectify} explicitly estimates the prediction uncertainty during training for filtering. CBST~\cite{zou2018unsupervised} employs category confidence for generating balanced pseudo labels. %A generative model-based refinement approach is proposed in~\cite{morerio2020generative}.
MetaCor~\cite{guo2021metacorrection} models the noise distribution of the pseudo-labels to enhance the generalization ability of the model on the target domain.
%\cite{zhang2022uda} trains an auxiliary network to minimize source domain bias.
ProDA~\cite{zhang2021prototypical} uses online-estimated class-wise feature centroids to rectify labels, by aligning soft prototypical assignments for different views of the same target. 

%We present a novel FFT-based strategy for training an auxiliary PRN module for effectively refining pseudo-labels. We believe through training our refinement module with perturbations of source and target logits based on style and semantics, the model learns to effectively refine noisy target predictions. This continuous learning process helps the improvement of the student model's representation using refined labels from the target domain. In contrast, prior works lack an explicit noise-handling strategy to address the domain gap, leading to potentially inferior models with low-quality pseudo-labels.

We present a novel FFT-based strategy for training an auxiliary PRN module to effectively refine pseudo-labels. By training the refinement module using perturbations in source and target logits, tied to style and semantics, our approach facilitates the adept refinement of noisy target predictions. This continuous learning process enhances the student model's representation using more precise labels from the target domain. In contrast, prior works lack an explicit noise-handling strategy to address domain gap, leading to potentially subpar models with low-quality pseudo-labels.
%Unlike prior studies, which lack a clear noise-handling strategy to tackle domain differences, our technique prevents the creation of subpar models linked to low-quality pseudo-labels.

%-------------------------------------------------------------------------
%-------------------------------------------------------------------------
\section{Methodology}

We first discuss the baseline self-training UDA method. Then, we discuss UDA with our proposed pseudo-label refinement neural network model and provide the details of our FFT-based perturbation approach to train the model. %that focuses on the refinement and localization of noise in pseudo-labels. 
Next, we discuss two additional components (i.e., contrastive learning, and Fourier-based style adaptation) of our framework which help to further improve the quality of our models. Finally, we discuss the overall loss function.

\vspace{0.1cm}
\noindent \textbf{Problem Setting:}
Let, $\mathsf{D}_\mathsf{S}=\{(x_\mathsf{S}^i,y_\mathsf{S}^i )\}_{i=1}^{N_\mathsf{s}}$ be the source domain dataset with $N_\mathsf{s}$ labeled samples where $y_\mathsf{S}^i$ denotes the one-hot ground-truth (GT) per pixel label for image $x_\mathsf{S}^i$. Here, $x_\mathsf{S}^i \in \mathbb R^{H\times W}$ and $y_\mathsf{S}^i \in \{0,1\}^{H\times W\times K}$. $(H, W)$ is the image resolution and $K$ is the number of classes. We also have a target domain dataset $\mathsf{D}_\mathsf{T}=\{(x_\mathsf{T}^i\}_{i=1}^{N_\mathsf{t}}$ containing $N_\mathsf{T}$ images without ground-truth labels. $\mathsf{D}_\mathsf{S}$ and $\mathsf{D}_\mathsf{T}$ share a common set of $K$ classes. In UDA semantic segmentation, the goal is to train a segmentation model $\mathcal{F}_{\theta}$ for the target domain by utilizing the labeled set $\mathsf{D}_\mathsf{S}$ from source domain and the unlabeled set $\mathsf{D}_\mathsf{T}$ from the target domain.

\subsection{Self-Training (ST) for UDA}
\label{sec:st_uda}
We can train a neural network model $\mathcal{F}_{\theta}$ with the available source domain images and labels employing supervised learning with a cross-entropy loss ${\mathcal{L}}^{S}_{ce}$  on source domain images. ${\mathcal{L}}^{S}_{ce}$ for $\text{i}^{\text{th}}$ sample can be written as,
\begin{equation}
    \vspace{-0.01cm}
    \mathcal{L}^{S(i)}_{ce} = - \sum_{j=1}^{H\times W} \sum_{k=1}^{K}{{y}_\mathsf{S}^{(i,j,k)} \log \mathcal{F}_{\theta}(x_\mathsf{S}^i)^{(j,k)}}
    \vspace{-1mm}
\end{equation}
However, due to the domain gap, the model trained with only source domain images with loss ${\mathcal{L}}^{S}_{ce}$ is unlikely to generalize well to the target domain. To address the domain gap, 
%several UDA techniques have been proposed (e.g., self-training, adversarial learning, cross-domain divergence minimization). 
we adopt the self-training-based UDA technique as our baseline
%, which has been shown to be stable and achieves state-of-the-art performance in UDA for semantic segmentation
~\cite{tranheden2021dacs,hoyer2022hrda,hoyer2022daformer,zhou2022context, bruggemann2023refign}. In self-training, a teacher model $\mathcal{F}_{\phi}$ is used for generating the pseudo-labels ${\hat{y}}_\mathsf{T}^{i}$ for target domain images. Pseudo-labeled target data is used along with labeled source data for training the student model $\mathcal{F}_{\theta}$ iteratively, to adapt the model to the target domain. In general, the semantic segmentation
models $\mathcal{F}_{\theta}$ ($\mathcal{E}_\theta$, $\mathcal{D}_\theta$) and  $\mathcal{F}_{\phi}$ ($\mathcal{E}_\phi$, $\mathcal{D}_\phi$) consist of a feature extractor (i.e., encoder $\mathcal{E}$), followed by a classifier predicting pixel-wise labels (i.e., decoder $\mathcal{D}$).

\begin{figure}[t]
    \centering
    \includegraphics[width=1\linewidth]{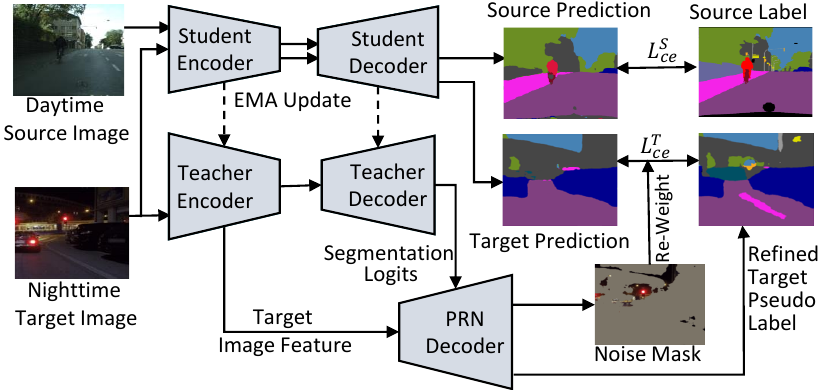}
    \caption{Overview of the UDA self-training framework with the proposed pseudo label refinement network (PRN). We consider the PRN decoder to be fixed (i.e., stop gradient flow to PRN) when calculating losses for training the student network.}
    \label{fig:framework}
\end{figure}

The teacher model is updated using the exponential moving average (EMA) of weights of the student model after each training step, which helps the teacher produce stable predictions. The teacher weights ${\phi}_{n+1}$ at train step $n+1$ is,
\begin{equation}
    {\phi}_{n+1} =  \beta {\phi}_{n} + (1-\beta) \theta_{n}, \label{ema}
\end{equation}
Here, $\beta$ is a hyper-parameter to adjust the degree of change in the model weights.
%during each iteration. 
The pseudo labels ${\hat{y}}_\mathsf{T}^{i}$ for target domain images are generated using the teacher model $\mathcal{F}_{\phi}$. 
\begin{equation}
    \vspace{-0.01cm}
    {\hat{y}}_\mathsf{T}^{i} = \argmax\limits_{k} \mathcal{F}_{\phi}(x_\mathsf{T}^i)^{(j,k)}
    \vspace{-1mm}
\end{equation}
These pseudo labels are also used to calculate an additional cross-entropy loss ${\mathcal{L}}^{T}_{ce}$ to train the student model $\mathcal{F}_{\theta}$ to adapt to the target domain. To minimize the effect of label noise, the $\mathcal{L}^{T(i)}_{ce}$ loss for target domain samples is weighted with quality estimates of the pseudo labels \cite{tranheden2021dacs, hoyer2022daformer, hoyer2022hrda}.
\iffalse
\begin{equation}
    \vspace{-0.01cm}
    \mathcal{L}^{T(i)}_{ce} = - \sum_{j=1}^{H\times W} \sum_{k=1}^{K} {{\hat{y}}_\mathsf{T}^{(i,j,k)} \log \mathcal{F}_{\theta}(x_\mathsf{T}^i)^{(j,k)}}
    \vspace{-1mm}
\end{equation}
\fi
%The imminent presence of noisy labels in this pseudo-label-based self-training setting is very likely to lead to sub-optimal performance \cite{arpit2017closer}. In this regard, we follow the recent state-of-the-art self-training-based UDA semantic segmentation works to adopt two strategies \cite{hoyer2022daformer, hoyer2022hrda, vayyat2022cluda}. 
\begin{equation}
    \vspace{-0.01cm}
    \mathcal{L}^{T(i)}_{ce} = - \sum_{j=1}^{H\times W} \sum_{k=1}^{K}{\eta}_T^i {{\hat{y}}_\mathsf{T}^{(i,j,k)} \log \mathcal{F}_{\theta}(x_\mathsf{T}^i)^{(j,k)}}
    \label{eqn:LT}
\end{equation}
Here, ${\eta}_T^i$ ($0\leq{\eta}_T^i\leq 1 $) is a confidence estimate of pseudo-label ${\hat{y}}_\mathsf{T}^{i}$. The labels are not always correct for the target domain samples, and the modified $\mathcal{L}^{T(i)}_{ce}$ loss takes that into consideration. On the other hand,
%Hence, the cross-entropy loss for target domain samples $\mathcal{L}^{T(i)}_{ce}$ is modified by weighting pseudo labels with a confidence estimate. 
we are fully confident about the source domain labels (i.e., ${\eta}_s^i$=1) as they are ground-truth and hence, $\mathcal{L}^{S(i)}_{ce}$ does not require any modification. Following prior works~\cite{hoyer2022daformer,hoyer2022hrda}, ${\eta}_T^i$ can be calculated as the percentage of pixels in the image with maximum softmax probability exceeding a threshold $\tau_1$.
\begin{equation}
    {\eta}_T^i = \frac{\sum_{j=1}^{H\times W} \mathbbm{1}[\max\limits_{k} \mathcal{F}_{\phi}(x_\mathsf{T}^i)^{(j,k)} > \tau_1]} {H \times W}
    \label{eq:qual_estimate}
\end{equation}
We also use augmented target data in training, which has been shown to be effective in prior works~\cite{araslanov2021self,tranheden2021dacs, hoyer2022daformer}. Data augmentation helps to learn more domain-robust features and thus improves generalization performance to unseen data. We use color jitter, gaussian blur, and ClassMix~\cite{olsson2021classmix} as data augmentations following prior works~\cite{hoyer2022daformer,tranheden2021dacs}. Augmented target samples are used in training the student model, while the non-augmented target samples are used by the teacher model to generate the pseudo-labels.

Despite adopting the strategies discussed above, ST approaches often suffer from the risk of training models that generalize poorly due to error propagation from the memorization of noisy pseudo labels (i.e., confirmation bias towards errors). To alleviate this issue, we propose to train an auxiliary pseudo-label refinement network (PRN) model $f_{\sigma}$ that focuses on label refinement and label noise localization.

%As only source labels are available, the supervised categorical cross-entropy loss can only be calculated for the source predictions

\subsection{UDA with Proposed Pseudo Label Refinement}

An overview of the UDA self-training framework with the proposed pseudo label refinement network (PRN) is shown in Fig.~\ref{fig:framework}. PRN network is a decoder model $\mathcal{D}_{\sigma}$ that takes in target image features from the teacher encoder $\mathcal{E}_{\phi}(x_\mathsf{T}^i)$ and segmentation logits from the teacher decoder $\mathcal{F}_{\phi}(x_\mathsf{T}^i)$. It outputs refined labels ${\Bar{y}}_\mathsf{T}^{i}$  and noise masks ${\mu}_\mathsf{T}^{i}$. Here, ${\mu}_\mathsf{T}^{(i,j)}$ is 1
if the pseudo-label of pixel j is predicted to be noisy and 0 otherwise. Different from Eq.~\ref{eq:qual_estimate}, we use noise mask ${\mu}_\mathsf{T}^{i}$ to calculate the confidence estimate ${\bar{\eta}}_T^i$.
\begin{equation}
    {\bar{\eta}}_T^i = \frac{\sum_{j=1}^{H\times W} \mathbbm{1}[{\mu}_\mathsf{T}^{(i,j)}=0 ]} {H \times W}
    \label{eq:qual_estimate2}
\end{equation}
%Similar to Eq.~\ref{eq:qual_estimate}, the PRN model segmentation logits output is used to calculate the confidence estimate ${\bar{\eta}}_T^i$.
The use of noise mask instead of segmentation logits, allows us to avoid selecting a threshold (i.e., $\tau_1$ in Eq.~\ref{eq:qual_estimate}) for the calculation of quality estimate.
The target cross-entropy loss $\mathcal{L}^{T(i)}_{ce}$ is modified by using ${\Bar{y}}_\mathsf{T}^{i}$ and ${\bar{\eta}}_T^i$. The source cross-entropy loss $\mathcal{L}^{S(i)}_{ce}$ remains the same as the source labels are GT and do not require any refinement.
\iffalse
We use ${\Bar{y}}_\mathsf{T}^{i}$,  ${\mu}_\mathsf{T}^{i}$ to modify the loss $\mathcal{L}^{T(i)}_{ce}$ for target data (Eq.~\ref{eqn:LT}).
\begin{equation}
    \small
    \mathcal{L}^{T(i)}_{ce} = -\sum_{j=1}^{H\times W} \sum_{k=1}^{K} \mathbbm{1}\{{\mu}_\mathsf{T}^{(i,j)}=0 \} ~{\bar{\eta}}_T^i {{\Bar{y}}_\mathsf{T}^{(i,j,k)} \log \mathcal{F}_{\theta}(x_\mathsf{T}^i)^{(j,k)}}
    \label{eqn:LT2}
\end{equation}
Here, $\mathcal{L}^{T(i)}_{ce}$ calculation avoids the difficult pixels for which the pseudo-label is predicted to be noisy. 
\fi
Based on the predicted target noise mask ${\mu}_\mathsf{T}^{i}$, $\mathcal{L}^{T(i)}_{ce}$ calculation can avoid the difficult pixels for which the pseudo-label is predicted to be noisy.
We use noise masks in creating the augmented target samples used for training the student model. For predicted difficult pixels in a target image, a randomly selected source image from the batch and corresponding GT labels are used in place of the target image pixels and pseudo labels. Next, we discuss how we train the refinement network in Sec.~\ref{train_refine}.

\subsection{Training the PRN Network}
\label{train_refine}

\begin{figure}[t]
    \centering
    \includegraphics[width=\linewidth]{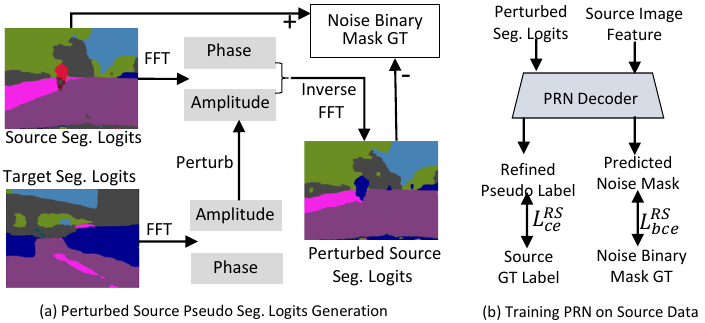}
    \caption{Perturbed label generation and training of PRN model on source data. A similar process is followed for target data as described in Sec.~\ref{train_refine}. In (a), we show a segmentation map instead of segmentation logits only for visualization purposes. We consider the student network to be fixed when training the PRN decoder.}
    \label{fig:PRN}
\end{figure}

The pseudo-label refinement network takes in source/target image features as well as perturbed segmentation logits and focuses on learning to predict higher-quality segmentation labels and noise max predicting pixels for which pseudo labels are likely to be erroneous. The model can be trained only using the labeled source data via supervised learning. However, such a model is unlikely to learn to effectively refine the target image pseudo labels. We train the PRN model using both source and target domain data using a novel FFT-based perturbation strategy. Prior work~\cite{yang2020fda} showed FFT-based transfer to be effective as a pre-processing step for UDA techniques. However, we consider FFT-based perturbation as an integral part of the UDA self-training process. 

First, we discuss how we perturb segmentation logits and generate pseudo noise masks for calculating losses on the source domain for training PRN. Fig.~\ref{fig:PRN} provides a brief illustration of the generation of perturbed source segmentation logits and PRN on source data. For source data, we have the ground-truth (GT) source label available and we use GT labels and predicted logits from the student network to calculate the cross-entropy loss. We perturb the source image segmentation logits using segmentation logits (from the teacher network) of a randomly sampled target domain image in batch. Based on the Fast Fourier Transform, the low-level frequencies of the amplitude of source segmentation logits are replaced by that of the target domain image. The perturbed segmentation logits are reconstituted using modified amplitude and unaltered phase via the inverse FFT (iFFT). Let's assume, $l_\mathsf{S}^i$ and $l_\mathsf{T}^i$ are segmentation logits from sampled source and target images. $\mathcal{T}^A$ and $\mathcal{T}^P$ denote the amplitude and phase components of the Fourier transform $\mathcal{T}$ of segmentation logits $l_\mathsf{S}^i$. The perturbed logits $\tilde{l}_\mathsf{S}^i$ calculation can be formalized as,
\begin{equation}
    \tilde{l}_\mathsf{S}^i = \mathcal{T}^{-1} ( [\mathsf{M}_{\epsilon} \cdot \mathcal{T}^A(l_\mathsf{T}^i)+ (1-\mathsf{M}_{\epsilon}) \cdot \mathcal{T}^A(l_\mathsf{S}^i), \mathcal{T}^P(l_\mathsf{S}^i)])
    \label{eq:ifft_source}
\end{equation}
$\mathcal{T}^{-1}$ denotes iFFT. $\mathsf{M}_{\epsilon}$ is a mask which is calculated as, 
 \begin{equation}
    \mathsf{M}_{\epsilon}(h,w) = \mathbbm{1}_{(h,w)\in[-\epsilon H:\epsilon H, -\epsilon W: \epsilon W]}
\end{equation}
To perturb the low-frequency component, the value of $\mathsf{M}_{\epsilon}$ is set to 1 only for the center region and 0 otherwise. $\epsilon$ is a hyper-parameter ($\epsilon \in (0,1)$) that controls perturbation strength. In our experiment, we randomly select the $\epsilon$ value between $0.05$ and $0.2$. As the low-level spectrum (amplitude) encodes style characteristics and the phase encodes high-level semantics, the source logits are perturbed significantly without affecting the high-level semantics. By refining source segmentation logits perturbed in this process, we hypothesize our PRN model will learn to refine some characteristics of target pseudo-label noise. 

After applying the FFT-based perturbation discussed above, the image's semantic content remains the same and we can use the available GT source label for training. The binary mask GT ${\mu}_{\mathsf{S}}^i$ can be created by comparing the original and perturbed logits (transformed to labels using $\argmax$): Set to 1 for identical labels and 0 for others. 
%The binary mask is then generated comparing the labels: Set to 1 for identical labels and 0 for others.
PRN uses this perturbed source segmentation logits and source student encoder feature to generate refined segmentation logits $\bar{l}_\mathsf{S}^i$ and noise masks $\bar{\nu}_\mathsf{S}^i$. Let, $ (\bar{l}_\mathsf{S}^i, \bar{\nu}_\mathsf{S}^i) = \mathcal{D}_{\sigma}(\mathcal{E}_{\theta}(x_\mathsf{S}^i), \tilde{l}_\mathsf{S}^i)$. To refine source labels, PRN is trained with two loss components (i.e., (1) cross-entropy loss $\mathcal{L}^{RS}_{ce}$ between refined segmentation label $\bar{l}_\mathsf{S}^i$ with GT source label ${y}_{\mathsf{S}}^i$ and (2) binary cross-entropy loss $\mathcal{L}^{RS}_{bce}$ between predicted noise mask $\bar{\nu}_\mathsf{S}^i$ 
 and GT binary noise mask ${\mu}_{\mathsf{S}}^i$). 
 \begin{equation}
    \vspace{-0.01cm}
    \mathcal{L}^{RS(i)}_{ce} = - \sum_{j=1}^{H\times W} \sum_{k=1}^{K}{{y}_\mathsf{S}^{(i,j,k)} \log (\bar{l}_\mathsf{S}^i)^{(j,k)}}
    \vspace{-1mm}
\end{equation}
 \begin{equation}
 \begin{split}
    \mathcal{L}^{RS(i)}_{bce} = - \sum_{j=1}^{H\times W} \Big({{\mu}_\mathsf{s}^{(i,j)} \log (\bar{\nu}_\mathsf{S}^i)^{(j)}} \\ + {(1-{\mu}_\mathsf{s}^{(i,j)}) \log (1 -\bar{\nu}_\mathsf{S}^i)^{(j)}} \Big) 
    \end{split}
\end{equation}
\iffalse
 \begin{equation}
    \vspace{-0.01cm}
    \mathcal{L}^{RS(i)}_{bce} = - \sum_{j=1}^{H\times W} \sum_{k\in(0,1)}{{\mu}_\mathsf{s}^{(i,j,k)} \log (\bar{\nu}_\mathsf{S}^i)^{(j,k)}}
    \vspace{-1mm}
\end{equation}
 \begin{equation}
    \mathcal{L}^{RS(i)}_{bce} = - \sum_{j=1}^{H\times W} (1-{{\mu}_\mathsf{S}^{(i,j)}) \log (1 - (\bar{\nu}_\mathsf{S}^i)^{(j)}} )
\end{equation}
 \begin{equation}
 \small
    \mathcal{L}^{RS(i)}_{bce} = - \sum_{j=1}^{H\times W}   \left( {{\mu}_\mathsf{S}^{(i,j)} \log (\bar{\nu}_\mathsf{S}^i)^{(j)}} + (1-{{\mu}_\mathsf{S}^{(i,j)}) \log (1 - (\bar{\nu}_\mathsf{S}^i)^{(j)}} ) \right)
\end{equation}
\fi
Next, we discuss the losses for training PRN on target domain data. For target data, we do not have access to ground-truth labels. For reference, we use pseudo-label generated from the refinement decoder with unperturbed target logits from the teacher model. Since the target pseudo label at the early stage of training can be noisy, we set a high threshold $\tau_{2}$ on selecting the pseudo labels. After the learning rate warm-up period is over, we assume the model to have some ability to localize the noise in the pseudo labels. Then, we avoid the noisy parts of the target pseudo-label based on the predicted error mask ${\mu}_\mathsf{T}^{i}$ when calculating the cross-entropy loss for the target. Similar to Eq.~\ref{eq:ifft_source}, the perturbed target segmentation logits  $\tilde{l}_\mathsf{T}^i$ are constituted by replacing low-frequency part of its amplitude by the amplitude of segmentation logits of a source image randomly sampled from the batch. $\tilde{l}_\mathsf{T}^i$ can be written as,
\begin{equation}
    \tilde{l}_\mathsf{T}^i = \mathcal{T}^{-1} ( [\mathsf{M}_{\epsilon} \cdot \mathcal{T}^A(l_\mathsf{S}^i)+ (1-\mathsf{M}_{\epsilon}) \cdot \mathcal{T}^A(l_\mathsf{T}^i), \mathcal{T}^P(l_\mathsf{T}^i)])
    \label{eq:ifft_target}
\end{equation}

PRN network takes in perturbed target logits and target feature from the teacher encoder, to output refined target logits $\bar{l}_\mathsf{T}^i$ and noise mask $\bar{\nu}_\mathsf{T}^i$. Here, $ (\bar{l}_\mathsf{T}^i, \bar{\nu}_\mathsf{T}^i) = \mathcal{D}_{\sigma}(\mathcal{E}_{\theta}(x_\mathsf{T}^i), \tilde{l}_\mathsf{T}^i)$. The binary mask ground truth ${\mu}_{\mathsf{T}}^i$is again created based on differences between the original and perturbed logits. Now, the cross-entropy loss $\mathcal{L}^{RT}_{ce}$ and binary cross-entropy loss $\mathcal{L}^{RT}_{bce}$ for training PRN for target data refinement and error localization can be written as, 
 \begin{equation}
    \vspace{-0.01cm}
    \mathcal{L}^{RT(i)}_{ce} = - \sum_{j=1}^{H\times W} \sum_{k=1}^{K} \mathbbm{1} [{\mu}_\mathsf{T}^{(i,j)}=0 ] ~{{\bar{y}}_\mathsf{T}^{(i,j,k)} \log (\bar{l}_\mathsf{T}^i)^{(j,k)}}
    \vspace{-1mm}
\end{equation}
 \begin{equation}
 \begin{split}
    \mathcal{L}^{RT(i)}_{bce} = - \sum_{j=1}^{H\times W}   \Big( {{\mu}_\mathsf{T}^{(i,j)} \log (\bar{\nu}_\mathsf{T}^i)^{(j)}} \\ + (1-{{\mu}_\mathsf{T}^{(i,j)}) \log (1 - (\bar{\nu}_\mathsf{T}^i)^{(j)}} ) \Big)
    \end{split}
\end{equation}
%Please note that, these losses used for training refinement decoder does not affect the  student networks, i.e., we consider student and teacher network fixed 

The student and refinement networks are trained simultaneously, but we consider one network fixed when calculating loss for the other. We discuss training loss in Sec.~\ref{sec:overall_loss}.

\subsection{Contrastive Learning and Fourier Adaptation}

%The imminent presence of noisy labels in this pseudo-label-based self-training setting is very likely to lead to sub-optimal performance \cite{arpit2017closer}. In this regard, 

To further stabilize the adaptation performance of model, we take two additional measures, i.e., pixel-wise contrastive loss (CL), and Fourier based style adaptation (FA).

We add the pixel-pixel contrastive loss $\mathcal{L}_{con}$ in training our student network. We hypothesize this addition will complement the source and target cross-entropy losses, for further improving the quality of our learned representations. Pixel-pixel contrastive loss has been shown in prior works to improve the training of semantic segmentation models ~\cite{wang2021exploring}. $\mathcal{L}_{con}$ attempts to pull the features of pixels of the same object class (i.e., positive pairs) close and push away the features of pixels of different object classes (i.e., negative pairs). We randomly select pairs of source and test domain samples from the input batch to calculate the loss. For a pixel $a$, let $\mathbf{X}_{\mathsf{P}}^a$ denote the set of all positive samples (i.e., pixel collection belonging to the same class of pixel $a$). Similarly, let $\mathbf{X}_{\mathsf{N}}^a$ denote the set of all negative samples (i.e., pixels not belonging to the same class of pixel $a$).
\begin{equation}
\small
    \mathcal{L}_{con} = -\frac{1}{|\mathbf{X}_{\mathsf{P}}^a|} \sum_{a^+\in \mathbf{X}_{\mathsf{P}}^a}\log \frac{\textbf{s}(f_a, f_{a^+})} {\textbf{s}(f_a, f_{a^+}))+\sum_{a^-\in \mathbf{X}_{\mathsf{N}}^a} \textbf{s}(f_a, f_{a^-})} 
\end{equation}
Here similarity function, $\textbf{s}(f_a, f_{a^+})$ is calculated as $\exp(\cos(f_a, f_{a^+})/\zeta)$. $\zeta$ is the temperature hyper-parameter that controls the similarity magnitude. We utilize GT labels for source samples and refined labels from the PRN network for target samples to find the positive and negative samples. The noise binary mask is used to avoid the target image pixels predicted to have incorrect labels. 

%\textcolor{red}{Incomplete: Fourier domain adaptation: Generate synthetic source image with target style to reduce appearance gap.  Motivate from use of paired images to reduce domain gap in low-light domain adaptation. This module does not require any learning, and can be easily integrated into our pipeline with a minimal added computational load. We add the FDA module to transfer style from target to source image based on fourier transform.}
%We empirically find this module to be especially helpful to compare with state-of-the-art approaches in low-light domain adaptation where most SOTA approaches use paired data collected across different time of day to boost performance. 

We adopt Fourier adaptation (FA) module following \cite{yang2020fda} to generate a synthetic source image with a target image style (without changing semantic content). As this module does not require any learning, it can be easily integrated into our pipeline with minimal additional load. The approach is similar to how we performed perturbation of logits (Eq.~\ref{eq:ifft_source}). However, we now focus on transforming the source image to the target style to reduce the perceptual gap between domains. These generated synthetic source images are then used in training instead of the original source images. 

\subsection{Overall Loss}
\label{sec:overall_loss}
The student network and PRN network are trained simultaneously. However, we consider the PRN network $\mathcal{F}_{\sigma}$ fixed when training the student network $\mathcal{F}_{\theta}$, i.e., the losses used for training the student network do not affect the PRN network. In this regard, we stop the gradient from flowing back in the
PRN network. Similarly, we train the PRN network $\mathcal{F}_{\sigma}$ considering the student network $\mathcal{F}_{\theta}$ is fixed. Finally, the overall optimization problem can be written as follows,
\begin{equation}
\small
    \min\limits_{\theta} (\mathcal{L}^{T}_{ce} + \mathcal{L}^{S}_{ce} + \lambda_{1}\mathcal{L}_{con}) + \min\limits_{\sigma} (\lambda_{2}(\mathcal{L}^{RS}_{ce}+ \mathcal{L}^{RS}_{bce})+\mathcal{L}^{RT}_{ce} +\mathcal{L}^{RT}_{bce} )
\end{equation}
Here, $\lambda_{1}$ and $\lambda_{2}$ are loss weight coefficients.

% ------------------------------------

\renewcommand{\arraystretch}{1.15}
\begin{table*}[]
\caption{Evaluation on \textbf{GTA$\to$Cityscapes}. We report mean IoU (mIoU) over 19 categories on the Cityscapes validations set.}
\vspace{-1.5mm}
\resizebox{1\linewidth}{!}{
\setlength{\tabcolsep}{3pt}
\begin{tabular}{c|c|ccccccccccccccccccc|c}
\toprule
Method    &    & Road  & S.Walk & Build.        & Wall  & Fence & Pole  & T.Light        & Sign  & Veget.        & Terrain       & Sky   & Person        & Rider & Car   & Truck & Bus   & Train & M.Bike & Bike  & \cellcolor{green!5}mIoU  \\
\midrule
CBST ~\cite{zou2018unsupervised}  & \parbox[t]{3mm}{\multirow{7}{*}{\rotatebox[origin=c]{90}{ResNet-Based}}} & 91.8  & 53.5  & 80.5  & 32.7  & 21.0  & 34.0  & 28.9  & 20.4  & 83.9  & 34.2  & 80.9  & 53.1  & 24.0  & 82.7  & 30.3  & 35.9  & 16.0  & 25.9  & 42.8  & \cellcolor{green!5}45.9  \\
CCM ~\cite{guanContent}  &  & 93.5 &57.6 &84.6 &39.3 &24.1 &25.2 &35.0 &17.3 &85.0 &40.6 &86.5 &58.7 &28.7 &85.8 &49.0 &56.4 &5.4 &31.9 &43.2  &\cellcolor{green!5}49.9  \\
MetaCor ~\cite{guo2021metacorrection} &    & 92.8 &  58.1 &  86.2 &  39.7 &  33.1 &  36.3 &  42.0 &  38.6 &  85.5 &  37.8 &  87.6 &  62.8 &  31.7 &  84.8 &  35.7 &  50.3 &  2.0 &  36.8 &  48.0 &  \cellcolor{green!5}52.1 \\
DACS ~\cite{tranheden2021dacs} &     & 89.9  & 39.7  & 87.9  & 30.7  & 39.5  & 38.5  & 46.4  & 52.8  & 88.0  & 44.0  & 88.8  & 67.2  & 35.8  & 84.5  & 45.7  & 50.2  & 0.0   & 27.3  & 34.0  & \cellcolor{green!5}52.2  \\
UAPLR ~\cite{wang2021uncertainty}  &    & 90.5 &38.7 &86.5 &41.1 &32.9 &40.5 &48.2 &42.1 &86.5 &36.8 &84.2 &64.5 &38.1 &87.2 &34.8 &50.4 &0.2 &41.8 &54.6   & \cellcolor{green!5}52.6 \\
CorDA ~\cite{wang2021domain}  &    & 94.7  & 63.1  & 87.6  & 30.7  & 40.6  & 40.2  & 47.8  & 51.6  & 87.6  & 47.0  & 89.7  & 66.7  & 35.9  & 90.2  & 48.9  & 57.5  & 0.0   & 39.8  & 56.0  & \cellcolor{green!5}56.6  \\
ProDA ~\cite{zhang2021prototypical}       & &87.8  & 56.0  & 79.7  & 45.3  & 44.8  & 45.6  & 53.5  & 53.5  & 88.6  & 45.2  & 82.1  & 70.7  & 39.2  & 88.8  & 45.5  & 59.4  & 1.0   & 48.9  & 56.4  & \cellcolor{green!5}57.5  \\
\textbf{DACS (w/ PRN)}    & & \textbf{92.7}  & \textbf{48.6}  & \textbf{88.9}  & \textbf{43.2}  & \textbf{33.3}  & \textbf{43.8}  & \textbf{49.0}  & \textbf{38.0}  & \textbf{88.4}  & \textbf{44.0}  & \textbf{86.5}  & \textbf{70.1}  & \textbf{45.0}  & \textbf{90.0}  & \textbf{41.4}  & \textbf{50.6}  & \textbf{42.0}  & \textbf{45.3}  & \textbf{58.7}  & \cellcolor{green!5}\textbf{57.9}  \\
\midrule
DAFormer~\cite{hoyer2022daformer} & \parbox[t]{3mm}{\multirow{5}{*}{\rotatebox[origin=c]{90}{SegFormer}}} & 95.7  & 70.2  & 89.4  & 53.5  & 48.1  & 49.6  & 55.8  & 59.4  & 89.9  & 47.9  & 92.5  & 72.2  & 44.7  & 92.3  & 74.5  & 78.2  & 65.1  & 55.9  & 61.8  & \cellcolor{green!5}68.2  \\
MIC-DAFormer~\cite{hoyer2023mic} & & 96.7 &75.0 &90.0 &58.2 &50.4 &51.1 &56.7 &62.1 &90.2 &51.3 &92.9 &72.4 &47.1 &92.8 &78.9 &83.4 &75.6 &54.2 &62.6 
  & \cellcolor{green!5}70.6  \\
\textbf{Ours}   &  & \textbf{95.8} & \textbf{73.3} & \textbf{92.8} & \textbf{56.2} & \textbf{51.9} & \textbf{51.6} & \textbf{59.6} & \textbf{62.8} & \textbf{93.1} & \textbf{49.9} & \textbf{96.3} & \textbf{76.1} & \textbf{47.0} & \textbf{96.3} & \textbf{77.7} & \textbf{81.7} & \textbf{68.2} & \textbf{59.9} & \textbf{64.3} & \cellcolor{green!5}\textbf{71.3} \\
\cline{1-1}\cline{3-22}
DAFormer~(w/ HRDA)~\cite{hoyer2022hrda} & &96.4	&74.4	&91.0	&61.6	&51.5	&57.1	&63.9	&69.3	&91.3	&48.4	&94.2	&79.0	&52.9	&93.9	&84.1	&85.7	&75.9	&63.9	&67.5	& \cellcolor{green!5}73.8 \\

%Ours (SegFormer w/o CL \& FA) & 95.9  & 72.8  & 89.5  & 54.6  & 51.0  & 52.1  & 57.8  & 59.2  & 91.0  & 49.6  & 95.0  & 74.9  & 44.1  & 94.4  & 75.4  & 80.5  & 66.7  & 57.7  & 62.4  & \cellcolor{green!5}69.7  \\
%Ours (SegFormer w/o CL)       & 96.2  & 73.9  & 91.7  & 55.4  & 52.1  & 52.1  & 58.4  & 60.3  & 93.0  & 50.9  & 93.7  & 75.7  & 48.5  & 94.9  & 77.2  & 81.6  & 66.7  & 58.1  & 64.1  & \cellcolor{green!5}70.8  \\
%Ours (SegFormer w/o FA)       & 96.2  & 71.7  & 92.3  & 55.6  & 51.1  & 52.5  & 58.3  & 61.7  & 91.7  & 50.9  & 95.1  & 74.4  & 45.9  & 95.0  & 77.2  & 80.9  & 67.7  & 57.8  & 64.5  & \cellcolor{green!5}70.5  \\
\textbf{Ours (w/ HRDA)}     &  & \textbf{96.4}	&\textbf{76.2}	&\textbf{90.9}	&\textbf{66.6}	&\textbf{53.6}	&\textbf{58.9}	&\textbf{63.3}	&\textbf{68.9}	&\textbf{92.3}	&\textbf{50.4}	&\textbf{95.2}	&\textbf{78.3}	&\textbf{54.8}	&\textbf{95.3}	&\textbf{84.8}	&\textbf{87.4}	&\textbf{74.7}	&\textbf{65.3}	&\textbf{70.8}
  & \cellcolor{green!5}\textbf{75.0}  \\
\bottomrule       
\end{tabular}
}
\label{tab:gta}
\end{table*}
\renewcommand{\arraystretch}{1}

%-------------------------------------------------------------------------
\section{Experiments}
\subsection{Experimental Setup}

\textbf{Datasets and Metrics.} The proposed method is evaluated on several UDA semantic segmentation benchmarks, i.e., CityScapes→Dark Zurich, GTA→Cityscapes,  SYNTHIA→Cityscapes.
%one normal-to-adverse condition and two synthetic-to-real unsupervised domain adaptation tasks. For normal-to-adverse condition adaptation, we consider CityScapes→Dark Zurich.  For synthetic-to-real adaptation, we consider GTA→Cityscapes, and  SYNTHIA→Cityscapes.
CityScapes (CS) contains daytime driving scenes from $50$ different cities~\cite{cordts2016cityscapes}.
%with resolution $2048 \times 1024$. 
Following prior works~\cite{hoyer2022daformer,yang2020fda,tranheden2021dacs}, we use $2,975$ training and $500$ validation images. Dark Zurich~\cite{sakaridis2019guided} is another driving dataset containing $8,779$ images (with GPS) captured at nighttime, twilight, and daytime with a resolution of $1080$p. Dark Zurich (DarkZ) contains $2,416$ unlabeled nighttime images for training. It also contains $201$ labeled nighttime images ($50$ validation, and $151$ test) for evaluation. The evaluation images have pixel-level annotations for the 19 classes of Cityscapes. 
%Images are collected using  1080p GoPro Hero 5 camera,
GTA~\cite{richter2016playing} is a synthetic dataset containing $24,966$ images with resolution $1914 \times 1052$ collected from GTA5 video game. The 19 classes common with CityScapes are used by SOTA methods for evaluation.
SYNTHIA (SYN) is another synthetic dataset with $9,400$ images with a resolution of $1280 \times 760$ ~\cite{ros2016synthia}. The 16 classes common with CityScapes are used for evaluation. 

\vspace{1mm}
\textbf{Network Architecture.}
Our default semantic segmentation network model is based on the SegFormer architecture~\cite{xie2021segformer} following recent state-of-the-art works~\cite{hoyer2022daformer,bruggemann2023refign}. The model consists of Transformer based MiT-B5 encoder (that generates hierarchical feature representation) and an MLP decoder (that aggregates information from multiple layers)~\cite{xie2021segformer}. We also train baselines with a ResNet101-based DeepLabV2 model. The Refine decoder utilizes the same MLP decoder architecture as described above. It utilizes aggregated information from the encoder output and segmentation logits. The number of channels in the input layers is increased to enable channel-wise concatenation of the two.
%both the DeepLabV2 decoder architecture from the DeepLabV2 model and the DAFormer decoder architecture from the DAFormer model. 
It includes two output layers: one for predicting segmentation maps and another for binary noise masks.

%-------------------------------------------------------------------------
\vspace{1mm}
\textbf{Implementation Details and Metrics.}
We follow DAFormer~\cite{hoyer2022daformer} training parameters in training our models. Our model is trained using AdamW. A learning rate of $6 \times 10^{-5}$ is used for the encoder. A learning rate of $6 \times 10^{-4}$  is used for both the student decoder and refinement decoder. The AdamW betas are set to $(0.9, 0.999)$, and weight decay is set to $0.01$. We use a batch size of 2, crop of $512$X$512$ and train for $40K$ iterations. We use linear learning rate warmup with a warmup rate of $10^{-6}$ for the first $1.5K$ iterations. The EMA weight parameter $\beta$ is set to $0.999$. $\tau_2$ is set to $0.968$. The loss weights $\lambda_1$ is set to $0.1$ and $\lambda_2$ is set to $25$. We set mask parameter $\epsilon$ to $0.005$ for the FA module. As the metric for semantic segmentation performance, we use mean intersection over union (mIoU). The results of our method are reported averaging over 3 random seeds.

\vspace{0.5mm}
\textbf{Computational Load.}
%The computation increase due to FFT and iFFT is minimal. In our test, using them resulted in about 4.9\% increase in computation time compared to without it.
We find our Framework increases training time by about 24.7$\%$ compared to the standard self-training (Sec. 3.2) baseline, DAFormer. Please note that the increase is only in training, while the inference time remains the same. The computation increase due to applying FFT is minimal (i.e., about 4.9\% increase in computation time compared to without it). Overall, we find the training of our model was completed in a reasonable time with limited resources (e.g., using a single 2080Ti GPU in about 19 hours for CityScapes→Dark Zurich). 

%In our test,
%, with time per iteration of 1.329 sec. and 1.035 sec. respectively. 
%Please note that the increase is only in training, and we can train in a reasonable time with limited resources (e.g., using a single 2080Ti GPU in about 19 hours for CityScapes→Dark Zurich). 

%[Need to add $\lambda_1$ and $\lambda_2$ values]

%and set α=0:99 and τ=0:968. 

%-------------------------------------------------------------------------
%\subsection{Baseline Methods}
%We consider a number of baselines and state-of-the-art methods for unsupervised domain adaptation.

% ------------------------------

\renewcommand{\arraystretch}{1.15}
\begin{table*}[]
\caption{Evaluation on \textbf{Cityscapes$\to$Dark-Zurich}. We report mean IoU (mIoU) over 19 common categories between these datasets.}
\vspace{-1.5mm}
\resizebox{\linewidth}{!}{
\setlength{\tabcolsep}{3pt}
\begin{tabular}{c|c|c|ccccccccccccccccccc|c}
\toprule
Method  &    & Ref. & Road & S.Walk & Build. & Wall & Fence & Pole & T.Light & Sign & Veget. & Terrain & Sky  & Person & Rider & Car  & Truck & Bus  & Train & M.Bike & Bike & \cellcolor{green!5}mIoU \\
\midrule
Source-Only ~\cite{chen2017deeplab}    & \parbox[t]{3mm}{\multirow{13}{*}{\rotatebox[origin=c]{90}{ResNet-Based}}} &x   & 79.0 & 21.8    & 53.0   & 13.3 & 11.2  & 22.5 & 20.2      & 22.1 & 43.5   & 10.4    & 18.0 & 37.4   & 33.8  & 64.1 & 6.4   & 0.0  & 52.3  & 30.4   & 7.4  & \cellcolor{green!5}28.8 \\
AdaptSegNet ~\cite{tsai2018learning}  & &x   & 86.1 & 44.2    & 55.1   & 22.2 & 4.8   & 21.1 & 5.6       & 16.7 & 37.2   & 8.4     & 1.2  & 35.9   & 26.7  & 68.2 & 45.1  & 0.0  & 50.1  & 33.9   & 15.6 & \cellcolor{green!5}30.4 \\
ADVENT ~\cite{vu2019advent}      & &x   & 85.8 & 37.9    & 55.5   & 27.7 & 14.5  & 23.1 & 14.0      & 21.1 & 32.1   & 8.7     & 2.0  & 39.9   & 16.6  & 64.0 & 13.8  & 0.0  & 58.8  & 28.5   & 20.7 & \cellcolor{green!5}29.7 \\
BDL ~\cite{li2019bidirectional}    & &x   & 85.3 & 41.1    & 61.9   & 32.7 & 17.4  & 20.6 & 11.4      & 21.3 & 29.4   & 8.9     & 1.1  & 37.4   & 22.1  & 63.2 & 28.2  & 0.0  & 47.7  & 39.4   & 15.7 & \cellcolor{green!5}30.8 \\
DACS ~\cite{tranheden2021dacs}    & &x   & 83.1 & 49.1    & 67.4   & 33.2 & 16.6  & 42.9 & 20.7      & 35.6 & 31.7   & 5.1     & 6.5  & 41.7   & 18.2  & 68.8 & 76.4  & 0.0  & 61.6  & 27.7   & 10.7 & \cellcolor{green!5}36.7 \\
\textbf{DACS (w/ PRN)} & &x   & \textbf{75.8} & \textbf{43.1}    & \textbf{54.5}   & \textbf{16.6} & \textbf{15.0}  & \textbf{36.2} & \textbf{35.9}      & \textbf{38.1} & \textbf{59.0}   & \textbf{28.6}    & \textbf{26.4} & \textbf{52.4}   & \textbf{45.8}  & \textbf{68.7} & \textbf{34.3}  & \textbf{1.5}  & \textbf{47.0}  & \textbf{30.7}   & \textbf{16.5} & \cellcolor{green!5}\textbf{38.2} \\
\cline{1-1}\cline{3-23}
DMAda ~\cite{dai2018dark}    & & \checkmark   & 75.5 & 29.1    & 48.6   & 21.3 & 14.3  & 34.3 & 36.8      & 29.9 & 49.4   & 13.8    & 0.4  & 43.3   & 50.2  & 69.4 & 18.4  & 0.0  & 27.6  & 34.9   & 11.9 & \cellcolor{green!5}32.1 \\
%GCMA ~\cite{sakaridis2019guided}    & & \checkmark    & 81.7 & 46.9    & 58.8   & 22.0 & 20.0  & 41.2 & 40.5      & 41.6 & 64.8   & 31.0    & 32.1 & 53.5   & 47.5  & 75.5 & 39.2  & 0.0  & 49.6  & 30.7   & 21.0 & \cellcolor{green!5}42.0 \\
MGCDA ~\cite{sakaridis2020map}    & & \checkmark    & 80.3 & 49.3    & 66.2   & 7.8  & 11.0  & 41.4 & 38.9      & 39.0 & 64.1   & 18.0    & 55.8 & 52.1   & 53.5  & 74.7 & 66.0  & 0.0  & 37.5  & 29.1   & 22.7 & \cellcolor{green!5}42.5 \\
CDAda ~\cite{xu2021cdada}   & & \checkmark    & 90.5 & 60.6    & 67.9   & 37.0 & 19.3  & 42.9 & 36.4      & 35.3 & 66.9   & 24.4    & 79.8 & 45.4   & 42.9  & 70.8 & 51.7  & 0.0  & 29.7  & 27.7   & 26.2 & \cellcolor{green!5}45.0 \\
%DANNet~\cite{wu2021dannet}     & & \checkmark    & 90.4 & 60.1    & 71.0   & 33.6 & 22.9  & 30.6 & 34.3      & 33.7 & 70.5   & 31.8    & 80.2 & 45.7   & 41.6  & 67.4 & 16.8  & 0.0  & 73.0  & 31.6   & 22.9 & \cellcolor{green!5}45.2 \\
DANIA~\cite{wu2021one}    & & \checkmark    & 91.5 & 62.7    & 73.9   & 39.9 & 25.7  & 36.5 & 35.7      & 36.2 & 71.4   & 35.3    & 82.2 & 48.0   & 44.9  & 73.7 & 11.3  & 0.1  & 64.3  & 36.7   & 22.7 & \cellcolor{green!5}47.0 \\
CCDistill ~\cite{gao2022cross}     & & \checkmark    & 89.6 & 58.1    & 70.6   & 36.6 & 22.5  & 33.0 & 27.0      & 30.5 & 68.3   & 33.0    & 80.9 & 42.3   & 40.1  & 69.4 & 58.1  & 0.1  & 72.6  & 47.7   & 21.3 & \cellcolor{green!5}47.5 \\
\hline
Source-Only~\cite{xie2021segformer}     & \parbox[t]{3mm}{\multirow{7}{*}{\rotatebox[origin=c]{90}{SegFormer-Based}}} &x   & 84.2 & 39.2    & 60.2   & 33.3 & 6.7   & 35.9 & 33.7      & 32.1 & 49.1   & 20.7    & 11.0 & 51.5   & 46.0  & 73.1 & 10.8  & 0.6  & 73.9  & 28.1   & 23.3 & \cellcolor{green!5}37.5 \\
DAFormer~\cite{hoyer2022daformer}     & &x   & 93.5 & 65.5    & 73.3   & 39.4 & 19.2  & 53.3 & 44.1  & 44.0 & 59.5   & 34.5   & 66.6 & 53.4   & 52.7  & 82.1 & 52.7  & 9.5  & 89.3  & 50.5   & 38.5 & \cellcolor{green!5}53.8 \\
MIC-DAFormer~\cite{hoyer2023mic}     & &x   & 88.2 & 60.5 & 73.5 & 53.5 &23.8 &52.3 &44.6 &43.8 &68.6 &34.0 &58.1 &57.8 &48.2 &78.7 &58.0 &13.3 &91.2 &46.1 &42.9 &\cellcolor{green!5}54.6 \\
Refign ~\cite{bruggemann2023refign}    & &  \checkmark   & 91.8 & 65.0    & 80.9   & 37.9 & 25.8  & 56.2 & 45.2      & 51.0 & 78.7   & 31.0    & 88.9 & 58.8   & 52.9  & 77.8 & 51.8  & 6.1  & 90.8  & 40.2   & 37.1 & \cellcolor{green!5}56.2 \\
\textbf{Ours}    & &x   & \textbf{94.3} & \textbf{74.8} & \textbf{82.5} & \textbf{53.2} & \textbf{26.4} & \textbf{62.3} & \textbf{43.6} & \textbf{49.9} & \textbf{66.3} & \textbf{37.1} & \textbf{69.3} & \textbf{67.9} & \textbf{61.9} & \textbf{81.2} & \textbf{53.9} & \textbf{13.8} & \textbf{90.5} & \textbf{44.7} & \textbf{35.0} & \cellcolor{green!5}\textbf{58.4} \\
\cline{1-1}\cline{3-23}
%Ours (SegFormer w/o CL \& FA)   & &x   & 94.1 & 70.3    & 80.1   & 50.2 & 24.5  & 58.2 & 39.5      & 45.3 & 65.3   & 35.2    & 65.9 & 62.3   & 54.1  & 80.3 & 54.2  & 13.2 & 88.5  & 42.1   & 36.1 & \cellcolor{green!5}55.8 \\
%Ours (SegFormer w/o CL)       & &x   & 94.8 & 74.2    & 82.6   & 52.5 & 26.3  & 62.5 & 43.4      & 49.1 & 65.6   & 34.7    & 68.9 & 68.3   & 59.4  & 80.9 & 53.7  & 14.6 & 90.1  & 43.8   & 35.8 & \cellcolor{green!5}58.0 \\
%Ours (SegFormer w/o FA)  & &x   & 95.2 & 72.3    & 80.2   & 52.2 & 24.0  & 59.8 & 39.2      & 46.9 & 64.6   & 35.7    & 67.1 & 65.7   & 54.8  & 80.7 & 54.8  & 12.7 & 88.3  & 43.7   & 36.5 & \cellcolor{green!5}56.5\\
DAFormer~(w/ HRDA)~\cite{hoyer2022hrda}     & &x   & 90.4 &56.3 &72.0 &39.5 &19.5 &57.8 &52.7 &43.1 &59.3 &29.1 &70.5 &60.0 &58.6 &84.0 &75.5 &11.2 &90.5 &51.6 &40.9 &\cellcolor{green!5}55.9  \\
\textbf{Ours~(w/ HRDA)}     & &x   & \textbf{92.9}	&\textbf{55.8}	&\textbf{74.5}	&\textbf{40.2}	&\textbf{21.3}	&\textbf{61.9}	&\textbf{53.9}	&\textbf{45.4}	&\textbf{63.9}	&\textbf{35.6}	&\textbf{76.9}	&\textbf{63.2}	&\textbf{64.3}	&\textbf{89.3}	&\textbf{71.2}	&\textbf{14.4}	&\textbf{89.5}	&\textbf{52.8} &\textbf{47.3}	
&\cellcolor{green!5}\textbf{58.6}  \\
\bottomrule
\end{tabular}
}
\label{tab:darkzurich}
\end{table*}
\renewcommand{\arraystretch}{1}
%-------------------------------------------------------------------------
\subsection{Experimental Results}
We provide GTA→CS and CS→DarkZ quantitative results in this section. We also provide GTA→CS ablation study and some qualitative examples. The SYN→CS experiments and more qualitative examples are in the supplementary. Also, ablation studies (i.e., CS→DarkZ, SYN→CS, refinement loss weights) are in the supplementary.

\subsubsection{GTA$\to$Cityscapes Results}

In Table~\ref{tab:gta},  we compare the performance of our approach on GTA$\to$Cityscapes adaptation, against several state-of-the-art UDA semantic segmentation approaches and baselines. We divide the table into 3 parts to aid our study. 

\vspace{0.5mm}
\noindent \textbf{SOTA Performance. } From the second part of the table, we see our method outperforms the best-performing prior state-of-the-art methods (i.e., DAFormer, MIC) by significant margins. We see $+3.1\%$ absolute improvement compared to DAFormer and $+0.7\%$ absolute improvement compared to MIC in mIOU (i.e., $68.2\%$ with DAFormer and $70.6\%$ with MIC compared to $71.3\%$  with ours). Encouragingly, the mIoU improves consistently over most classes. We believe the masked image consistency module from MIC can be easily integrated into our method to further improve performance. We leave it as a future work.

\noindent  \textbf{Performance with HRDA Training. } We report the performance of our model with HRDA-based training in the last row of Table~\ref{tab:gta}, i.e., Ours (w/ HRDA). HRDA~\cite{hoyer2022hrda} is a recent UDA training approach that allows training with high-resolution images and has been shown to be able to boost UDA performance. 
%We mostly consider the performance without HRDA in our results analysis to enable fair comparison as most prior works don't use such a strategy. 
%using high-resolution images in training. 
We see incorporating HRDA training further improves our performance ($75.0\%$ vs. $71.3\%$). We also see the improvement is consistent with that of DAFormer with HRDA. Ours (w/ HRDA) outperforms  DaFormer (w/ HRDA) by $+1.2\%$ absolute mIoU.

\noindent  \textbf{Pseudo-Label Refinement Methods. }In the first part of Table~\ref{tab:gta}, we compared with several prior SOTA PL refinement methods for ResNet CNN-based UDA semantic segmentation (e.g., CBST [70],  CCM, MetaCor, UAPLR, ProDA). Among all the methods, we find incorporating our proposed PRN module with DACS, i.e., DACS (w/ PRN), performs the best. For example, Ours DACS (w/ PRN) achieves $+0.4\%$ improvement over ProDa and $+1.3\%$ over CorDA. Our explicit noise-handling approach with a learned auxiliary module helps us better address the domain gap compared to prior works, leading to potentially more resilient models with superior pseudo-labels.

%\textbf{Transformer vs CNN based Methods.}
%We find that Transformer-based UDA methods perform significantly better than CNN-based methods, which is expected. From 

%From the last part of Table~\ref{tab:gta}, we found that the performance drop is limited after removing one of these CL and FA components. Our method still outperforms the state-of-the-art DAFormer even after removing these two components.  

%HRDA allows training with high-resolution images for UDA. We didn't report using this strategy to enable fair comparison as most prior works don't use it [\textbf{L779 - L786}]. However, HRDA training is designed to be easily integrated with any self-training UDA techniques, such as Ours, DAFormer, etc. 

%contrast to most existing UDA semantic segmentation works that use low-resolution images. In this work, we follow the DAFormer~\cite{hoyer2022daformer} UDA approach and do not use high-resolution images. We compare performance with the methods that also utilize the same image resolutions.

%For example, DAFormer+HRDA showed a 5.5$\%$ increase in mIoU in GTA$\to$Cityscapes compared to baseline DAFormer. We also observe a similar degree of consistent improvement with our method, i.e., our best model achieves $6.1\%$ further improvement in mIoU in GTA$\to$Cityscapes incorporating HRDA training, outperforming SOTA DAFormer+HRDA by a large margin. %(74.3 vs. 73.8).

%-------------------------------------------------------------------------

\subsubsection{Cityscapes$\to$Dark Zurich Results}
We report the performance of our approach on Cityscapes$\to$Dark-Zurich adaptation in Table~\ref{tab:darkzurich}. We compare our approach with several state-of-the-art approaches. Most of the existing adverse-weather adaptation methods use additional reference images (depicting the same scene in the source domain) from the target domain, as auxiliary data to improve domain adaptation performance. Although these approaches show promising results, the requirement of collecting images of the same scene from both source and target domains, limits the applicability of these approaches. Note that our approach does not utilize any additional reference images, and still outperforms all these state-of-the-art approaches significantly. 

In the first and second parts of the table, we report several CNN-based methods. In the third and fourth parts of the table, we report transformer-based methods. As expected, CNN-based models perform worse than the Trasformer-based models. 
%We observe our DeepLab-v2-based baseline performs better than several other DeepLab-v2 baselines that do not use paired reference images. 
From table~\ref{tab:darkzurich}, the best-performing state-of-the-art method is Refign~\cite{bruggemann2023refign}, which also uses additional reference images to boost adaptation performance. Our method does not use any additional reference data, but still shows $+2.2\%$ absolute improvement in mIoU over Refign (i.e., $56.2\%$ with Refign compared to $58.4\%$ with ours). We also observe that our model performs the best in almost all the categories. Recent SOTA MIC-DAFormer~\cite{hoyer2023mic} performs best among the methods that do not use reference images. Our model achieves an absolute improvement of $+3.8\%$ over MIC-DAFormer (i.e., $54.6\%$ with MIC compared to $58.4\%$ with ours). In the last part of the table, we also report the performance of our method and DAFormer incorporating HRDA training. We again observe that our method with HRDA performs significantly better ($+2.7\%$) than DAFormer with HRDA.

\renewcommand{\arraystretch}{1}
\begin{table}[h]
\caption{Ablation study with different components of our proposed method on \textbf{GTA$\to$Cityscapes}.}
\vspace{-1mm}
\resizebox{1\linewidth}{!}{
\begin{tabular}{c|ccccccc|c}
\toprule
\# & ST & PL-R & NM & CL w/o R & CL w/ R & FA &HRDA & mIoU \\
\midrule
3.1 & x    & x    & x     & x    & x   & x & x  & 45.6 \\
3.2 & \checkmark    & x    & x     & x    & x   & x & x & 68.3 \\
3.3 & \checkmark    &  \checkmark    & x     & x    & x   & x & x & 69.7 \\
3.4 & \checkmark    &  \checkmark    &  \checkmark     & x    & x  & x & x  & 70.1 \\
3.5 & \checkmark    &  \checkmark    &  \checkmark     &  \checkmark    & x  & x & x  & 70.1 \\
3.6 & \checkmark    &  \checkmark    &  \checkmark     & x    &  \checkmark   & x & x & 70.5 \\
3.7 & \checkmark    &  \checkmark    &  \checkmark     & x    & x   &  \checkmark & x &  70.8   \\
\midrule
\rowcolor{aliceblue} \textbf{3.8} &  \checkmark    &  \checkmark    &  \checkmark     & x    &  \checkmark   &  \checkmark & x & \textbf{71.3} \\
\midrule
3.9 & \checkmark    &  \checkmark    &  \checkmark     &  x    & x  & x & \checkmark  & 74.3 \\
3.10 & \checkmark    &  \checkmark    &  \checkmark     & x    &  \checkmark   & x & \checkmark & 74.6 \\
3.11 & \checkmark    &  \checkmark    &  \checkmark     & x    & x   &  \checkmark & \checkmark &  74.8   \\
\midrule
\rowcolor{aliceblue} \textbf{3.12} &  \checkmark    &  \checkmark    &  \checkmark     & x    &  \checkmark   &  \checkmark & \checkmark & \textbf{75.0} \\
\bottomrule
\end{tabular}
}
\label{tab:ablation}
\end{table}
\renewcommand{\arraystretch}{1}

\subsubsection{Ablation Studies}

In Table~\ref{tab:ablation}, We perform an ablation study with different components of the method, i.e., Self-Training (ST), Pseudo Label Refinement (PL-R), Noise Mask (NM), Contrastive Learning without or with using the output of PRN (CL w/o R, CL w/ R),  Fourier Adaptation (FA) and HRDA training. 

\textbf{PRN Module: } Comparing rows 3.1 and 3.2 in Table~\ref{tab:ablation}, it is evident that the baseline self-training approach is effective in adapting from source to target domain (i.e., mIoU improves $+22.7\%$). Based on rows 3.2, and 3.4, we see that our pseudo-label refinement model leads to $+1.8\%$ absolute improvement in mIoU and hence, playing a vital role in achieving state-of-the-art accuracy. Comparing 3.3 and 3.4, we see that including noise mask (NM) prediction in PRN along with pseudo-label refinement (PL-R) leads to significant improvement in performance. 

\textbf{CL and FA: } Rows 3.5 and 3.6 indicate the performance change by adding contrastive learning (CL) module. Comparing row 3.4 with row 3.5, we see no performance change when the CL module does not use the PRN model output. However, comparing row 3.4 with row 3.6, we see a performance improvement of $0.4\%$ when PRN model output is used. Based on this, we find that our proposed pseudo-label refinement module is critical for the effective use of pixel-wise contrastive learning. It is because its success relies on the quality of pseudo labels to select positive and negative pairs. Based on row 3.7 and row 3.8, we see adding FA style adaptation module helps to improve adaptation performance. 
Overall, we see CL and FA components are complementary to our main contribution. With the proposed PRN, they show consistent improvement across benchmarks. 
%Specifically, CL is attached to our method,
%The two methods are complementary, 
%as evidenced by Table 4, where we observe a performance drop when CL (4.4 vs. 4.5) does not use our PRN output.

\textbf{Overall Framework:} The difference between rows 3.2 and 3.8 shows that our approach helps significantly to improve performance over the baseline self-training approach ($+3.0\%$ absolute improvement in mIoU). Going further, we also show how HRDA-based training can boost our UDA performance. Rows 3.9, 3.10, and 3.11 show the effect of using HRDA training for experiments 3.4, 3.6, and 3.7 respectively, and all of them show a consistent improvement in absolute mIoU. Finally, comparing row 3.12 and row 3.8, we can see that HRDA-training provides a significant boost of 3.7\% to our UDA pipeline. Please see the supplementary for more ablation studies.

%We have presented quantitative results in this section above. Please see the supplementary material for qualitative results.

%-------------------------------------------------------------------------

\subsubsection{Qualitative Results}

\begin{figure}[t]
    \centering
\includegraphics[width=\linewidth]{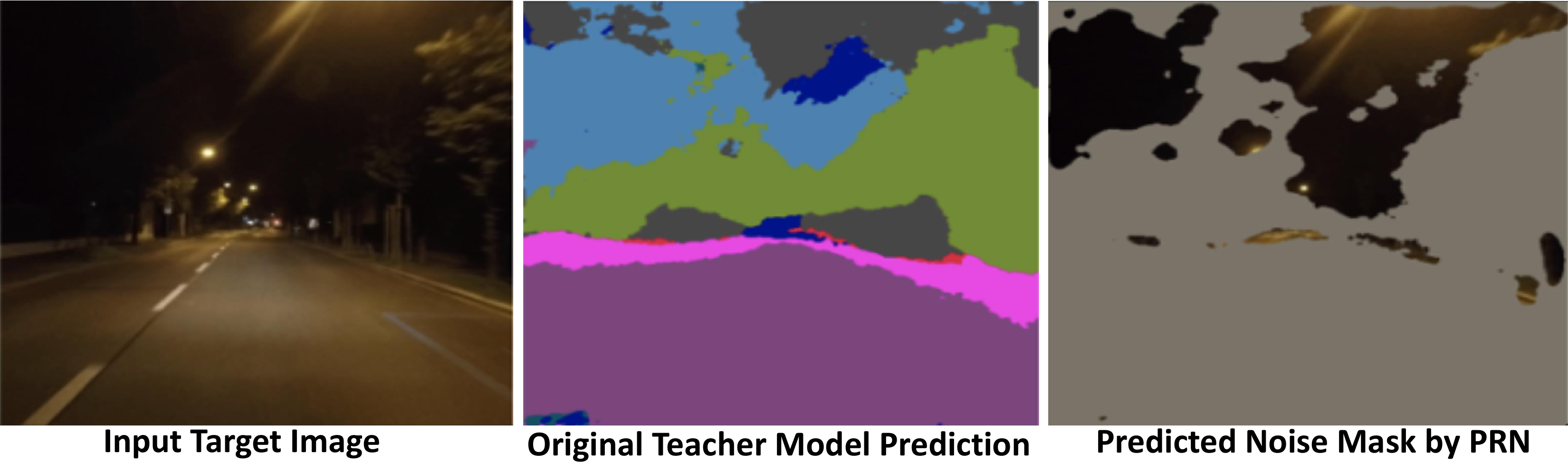}
\vspace{-0.6cm}
    \caption{An example to analyze PRN module prediction quality.}
    \label{fig:PRN_qual}
\end{figure}

\begin{figure}[t]
    \centering
\includegraphics[width=\linewidth]{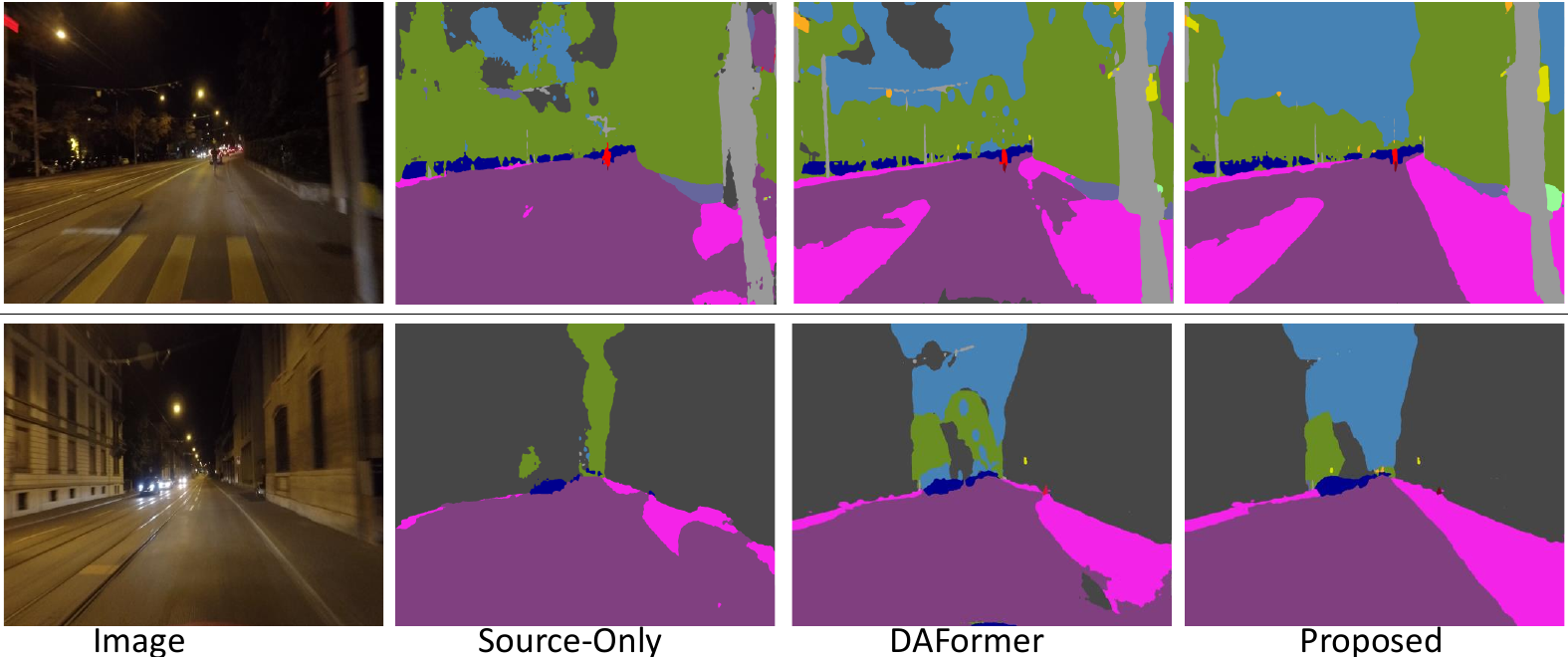}
    \vspace{-0.6cm}
    \caption{Qualitative comparison of the proposed on CS$\to$DarkZ}
    \label{fig:qualit}
\end{figure}

We provide an example in Fig.~\ref{fig:PRN_qual} showing teacher prediction and predicted noise mask to qualitatively evaluate the PRN module. Another example of the original teacher model prediction and noise mask is shown in Fig.~\ref{fig:framework}. We see noise mask generally corresponds well with the noisy part of pseudo labels in both cases. We also show two qualitative examples showing a semantic map generated from our model comparing source-only and DAFormer models in Fig.~\ref{fig:qualit}. We observe that the proposed method performs significantly better qualitatively than other approaches. Please see the supplementary for more qualitative results.

%------------------------------------------------------------------------
\section{Conclusion}
In this work, we present a novel self-training-based framework for the unsupervised adaptation of semantic segmentation models. We propose training auxiliary pseudo-label refinement network that helps self-training to be less susceptible to erroneous pseudo-label predictions by localizing and refining them.
We also introduce two additional components, contrastive learning and Fourier-based style adaptation in our framework to further improve the quality of the trained model. Our proposed approach shows significant performance improvement compared to the previous state-of-the-art approaches in both normal-to-adverse and synthetic-to-real adaptation benchmarks.

%%%%%%%%% REFERENCES
{\small
\bibliographystyle{ieee_fullname}
\bibliography{egbib}

\begin{thebibliography}{10}\itemsep=-1pt

\bibitem{araslanov2021self}
Nikita Araslanov and Stefan Roth.
\newblock Self-supervised augmentation consistency for adapting semantic segmentation.
\newblock In {\em Proceedings of the IEEE/CVF Conference on Computer Vision and Pattern Recognition}, pages 15384--15394, 2021.

\bibitem{arpit2017closer}
Devansh Arpit, Stanislaw Jastrzebski, Nicolas Ballas, David Krueger, Emmanuel Bengio, Maxinder~S Kanwal, Tegan Maharaj, Asja Fischer, Aaron Courville, Yoshua Bengio, et~al.
\newblock A closer look at memorization in deep networks.
\newblock In {\em International conference on machine learning}, pages 233--242. PMLR, 2017.

\bibitem{karstenMMD}
Karsten~M. Borgwardt, Arthur Gretton, Malte~J. Rasch, Hans-Peter Kriegel, Bernhard Sch¨olkopf, and Alex~J. Smola.
\newblock Integrating structured biological data by kernel maximum mean discrepancy.
\newblock {\em Bioinformatics}, 22:49--57, 2006.

\bibitem{bruggemann2023refign}
David Br{\"u}ggemann, Christos Sakaridis, Prune Truong, and Luc Van~Gool.
\newblock Refign: Align and refine for adaptation of semantic segmentation to adverse conditions.
\newblock In {\em Proceedings of the IEEE/CVF Winter Conference on Applications of Computer Vision}, pages 3174--3184, 2023.

\bibitem{chen2017deeplab}
Liang-Chieh Chen, George Papandreou, Iasonas Kokkinos, Kevin Murphy, and Alan~L Yuille.
\newblock Deeplab: Semantic image segmentation with deep convolutional nets, atrous convolution, and fully connected crfs.
\newblock {\em IEEE transactions on pattern analysis and machine intelligence}, 40(4):834--848, 2017.

\bibitem{choiSelfEns}
Jaehoon Choi, Taekyung Kim, and Changick Kim.
\newblock Self-ensembling with gan-based data augmentation for domain adaptation in semantic segmentation.
\newblock In {\em IEEE International Conference on Computer Vision (ICCV)}, 2019.

\bibitem{cordts2016cityscapes}
Marius Cordts, Mohamed Omran, Sebastian Ramos, Timo Rehfeld, Markus Enzweiler, Rodrigo Benenson, Uwe Franke, Stefan Roth, and Bernt Schiele.
\newblock The cityscapes dataset for semantic urban scene understanding.
\newblock In {\em Proceedings of the IEEE conference on computer vision and pattern recognition}, pages 3213--3223, 2016.

\bibitem{csurka2017comprehensive}
Gabriela Csurka.
\newblock A comprehensive survey on domain adaptation for visual applications.
\newblock {\em Domain adaptation in computer vision applications}, pages 1--35, 2017.

\bibitem{csurka2021unsupervised}
Gabriela Csurka, Riccardo Volpi, and Boris Chidlovskii.
\newblock Unsupervised domain adaptation for semantic image segmentation: a comprehensive survey.
\newblock {\em arXiv preprint arXiv:2112.03241}, 2021.

\bibitem{udaSurvey}
Gabriela Csurka, Riccardo Volpi, and Boris Chidlovskii.
\newblock Unsupervised domain adaptation for semantic image segmentation: a comprehensive survey.
\newblock In {\em arXiv:2112.03241}, 2021.

\bibitem{dai2018dark}
Dengxin Dai and Luc Van~Gool.
\newblock Dark model adaptation: Semantic image segmentation from daytime to nighttime.
\newblock In {\em 2018 21st International Conference on Intelligent Transportation Systems (ITSC)}, pages 3819--3824. IEEE, 2018.

\bibitem{gao2022cross}
Huan Gao, Jichang Guo, Guoli Wang, and Qian Zhang.
\newblock Cross-domain correlation distillation for unsupervised domain adaptation in nighttime semantic segmentation.
\newblock In {\em IEEE Conference on Computer Vision and Pattern Recognition (CVPR)}, pages 9913--9923, 2022.

\bibitem{guo2021metacorrection}
Xiaoqing Guo, Chen Yang, Baopu Li, and Yixuan Yuan.
\newblock Metacorrection: Domain-aware meta loss correction for unsupervised domain adaptation in semantic segmentation.
\newblock In {\em Proc. IEEE/CVF Conference on Computer Vision and Pattern Recognition}, pages 3927--3936, 2021.

\bibitem{adaptive_pyramid}
Junjun He, Zhongying Deng, Lei Zhou, Yali Wang, and Yu. Qiao.
\newblock Adaptive pyramid context network for semantic segmentation.
\newblock In {\em IEEE Conference on Computer Vision and Pattern Recognition (CVPR)}, 2019.

\bibitem{hoyer2022daformer}
Lukas Hoyer, Dengxin Dai, and Luc Van~Gool.
\newblock Daformer: Improving network architectures and training strategies for domain-adaptive semantic segmentation.
\newblock In {\em Proc. IEEE/CVF Conference on Computer Vision and Pattern Recognition}, pages 9924--9935, 2022.

\bibitem{hoyer2022hrda}
Lukas Hoyer, Dengxin Dai, and Luc Van~Gool.
\newblock Hrda: Context-aware high-resolution domain-adaptive semantic segmentation.
\newblock In {\em Proc. European Conference on Computer Vision (ECCV)}, pages 372--391. Springer, 2022.

\bibitem{hoyer2023mic}
Lukas Hoyer, Dengxin Dai, Haoran Wang, and Luc Van~Gool.
\newblock Mic: Masked image consistency for context-enhanced domain adaptation.
\newblock In {\em Proceedings of the IEEE/CVF Conference on Computer Vision and Pattern Recognition}, pages 11721--11732, 2023.

\bibitem{jiaxingAdv}
Jiaxing Huang, Dayan Guan, Aoran Xiao, and Shijian Lu.
\newblock Rda: Robust domain adaptation via fourier adversarial attacking.
\newblock In {\em IEEE International Conference on Computer Vision (ICCV)}, 2021.

\bibitem{karim2023c}
Nazmul Karim, Niluthpol~Chowdhury Mithun, Abhinav Rajvanshi, Han-pang Chiu, Supun Samarasekera, and Nazanin Rahnavard.
\newblock C-sfda: A curriculum learning aided self-training framework for efficient source free domain adaptation.
\newblock In {\em IEEE/CVF Conference on Computer Vision and Pattern Recognition}, pages 24120--24131, 2023.

\bibitem{ke2020guided}
Zhanghan Ke, Di Qiu, Kaican Li, Qiong Yan, and Rynson~WH Lau.
\newblock Guided collaborative training for pixel-wise semi-supervised learning.
\newblock In {\em Computer Vision--ECCV 2020: 16th European Conference, Glasgow, UK, August 23--28, 2020, Proceedings, Part XIII 16}, pages 429--445. Springer, 2020.

\bibitem{myeongTexture}
Myeongjin Kim and Hyeran Byun.
\newblock Learning texture invariant representation for domain adaptation of semantic segmentation.
\newblock In {\em IEEE Conference on Computer Vision and Pattern Recognition (CVPR)}, 2020.

\bibitem{kwon2022semi}
Donghyeon Kwon and Suha Kwak.
\newblock Semi-supervised semantic segmentation with error localization network.
\newblock In {\em Proceedings of the IEEE/CVF Conference on Computer Vision and Pattern Recognition}, pages 9957--9967, 2022.

\bibitem{guanContent}
Guangrui Li, Guoliang Kang, Wu Liu, Yunchao Wei, and Yi Yang.
\newblock Content-consistent matching for domain adaptive semantic segmentation.
\newblock In {\em Proc. European Conference on Computer Vision (ECCV)}, 2020.

\bibitem{li2019bidirectional}
Yunsheng Li, Lu Yuan, and Nuno Vasconcelos.
\newblock Bidirectional learning for domain adaptation of semantic segmentation.
\newblock In {\em IEEE Conference on Computer Vision and Pattern Recognition (CVPR)}, pages 6936--6945, 2019.

\bibitem{qingCurriculum}
Qing Lian, Fengmao Lv, Lixin Duan, and Boqing Gong.
\newblock Constructing self-motivated pyramid curriculums for cross-domain semantic segmentation: A non-adversarial approach.
\newblock In {\em IEEE International Conference on Computer Vision (ICCV)}, 2019.

\bibitem{mei2020instance}
Ke Mei, Chuang Zhu, Jiaqi Zou, and Shanghang Zhang.
\newblock Instance adaptive self-training for unsupervised domain adaptation.
\newblock In {\em European conference on computer vision}, pages 415--430. Springer, 2020.

\bibitem{mendel2020semi}
Robert Mendel, Luis~Antonio De~Souza, David Rauber, Joao~Paulo Papa, and Christoph Palm.
\newblock Semi-supervised segmentation based on error-correcting supervision.
\newblock In {\em Computer Vision--ECCV 2020: 16th European Conference, Glasgow, UK, August 23--28, 2020, Proceedings, Part XXIX 16}, pages 141--157. Springer, 2020.

\bibitem{olsson2021classmix}
Viktor Olsson, Wilhelm Tranheden, Juliano Pinto, and Lennart Svensson.
\newblock Classmix: Segmentation-based data augmentation for semi-supervised learning.
\newblock In {\em Proceedings of the IEEE/CVF Winter Conference on Applications of Computer Vision}, pages 1369--1378, 2021.

\bibitem{patel2015visual}
Vishal~M Patel, Raghuraman Gopalan, Ruonan Li, and Rama Chellappa.
\newblock Visual domain adaptation: A survey of recent advances.
\newblock {\em IEEE signal processing magazine}, 32(3):53--69, 2015.

\bibitem{chrisSelfEns}
Christian~S. Perone, Pedro Ballester, Rodrigo~C. Barros, , and Julien Cohen-Adad.
\newblock Unsupervised domain adaptation for medical imaging segmentation with self-ensembling.
\newblock {\em NeuroImage}, 194:1--11, 2019.

\bibitem{piotrowski1982demonstration}
Leon~N Piotrowski and Fergus~W Campbell.
\newblock A demonstration of the visual importance and flexibility of spatial-frequency amplitude and phase.
\newblock {\em Perception}, 11(3):337--346, 1982.

\bibitem{richter2016playing}
Stephan~R Richter, Vibhav Vineet, Stefan Roth, and Vladlen Koltun.
\newblock Playing for data: Ground truth from computer games.
\newblock In {\em European conference on computer vision}, pages 102--118. Springer, 2016.

\bibitem{ros2016synthia}
German Ros, Laura Sellart, Joanna Materzynska, David Vazquez, and Antonio~M Lopez.
\newblock The synthia dataset: A large collection of synthetic images for semantic segmentation of urban scenes.
\newblock In {\em Proceedings of the IEEE conference on computer vision and pattern recognition}, pages 3234--3243, 2016.

\bibitem{sakaridis2019guided}
Christos Sakaridis, Dengxin Dai, and Luc~Van Gool.
\newblock Guided curriculum model adaptation and uncertainty-aware evaluation for semantic nighttime image segmentation.
\newblock In {\em Proceedings of the IEEE/CVF International Conference on Computer Vision}, pages 7374--7383, 2019.

\bibitem{sakaridis2020map}
Christos Sakaridis, Dengxin Dai, and Luc Van~Gool.
\newblock Map-guided curriculum domain adaptation and uncertainty-aware evaluation for semantic nighttime image segmentation.
\newblock {\em IEEE Transactions on Pattern Analysis and Machine Intelligence}, 44(6):3139--3153, 2020.

\bibitem{robinSegmenter}
Robin Strudel, Ricardo Garcia, Ivan Laptev, and Cordelia Schmid.
\newblock Segmenter: Transformer for semantic segmentation.
\newblock In {\em IEEE International Conference on Computer Vision (ICCV)}, 2021.

\bibitem{subhani2020learning}
M~Naseer Subhani and Mohsen Ali.
\newblock Learning from scale-invariant examples for domain adaptation in semantic segmentation.
\newblock In {\em Computer Vision--ECCV 2020: 16th European Conference, Glasgow, UK, August 23--28, 2020, Proceedings, Part XXII 16}, pages 290--306. Springer, 2020.

\bibitem{tarvainen2017mean}
Antti Tarvainen and Harri Valpola.
\newblock Mean teachers are better role models: Weight-averaged consistency targets improve semi-supervised deep learning results.
\newblock {\em Advances in neural information processing systems}, 30, 2017.

\bibitem{tian2022striking}
Junjiao Tian, Niluthpol~Chowdhury Mithun, Zachary Seymour, Han-Pang Chiu, and Zsolt Kira.
\newblock Striking the right balance: Recall loss for semantic segmentation.
\newblock In {\em International Conference on Robotics and Automation (ICRA)}, pages 5063--5069. IEEE, 2022.

\bibitem{tranheden2021dacs}
Wilhelm Tranheden, Viktor Olsson, Juliano Pinto, and Lennart Svensson.
\newblock Dacs: Domain adaptation via cross-domain mixed sampling.
\newblock In {\em Proc. IEEE/CVF Winter Conference on Applications of Computer Vision}, pages 1379--1389, 2021.

\bibitem{tsai2018learning}
Yi-Hsuan Tsai, Wei-Chih Hung, Samuel Schulter, Kihyuk Sohn, Ming-Hsuan Yang, and Manmohan Chandraker.
\newblock Learning to adapt structured output space for semantic segmentation.
\newblock In {\em Proceedings of the IEEE conference on computer vision and pattern recognition}, pages 7472--7481, 2018.

\bibitem{vu2019advent}
Tuan-Hung Vu, Himalaya Jain, Maxime Bucher, Matthieu Cord, and Patrick P{\'e}rez.
\newblock Advent: Adversarial entropy minimization for domain adaptation in semantic segmentation.
\newblock In {\em Proceedings of the IEEE/CVF Conference on Computer Vision and Pattern Recognition}, pages 2517--2526, 2019.

\bibitem{wangSelfEns}
Kaihong Wang, Chenhongyi Yang, and Margrit Betke.
\newblock Consistency regularization with high-dimensional non-adversarial source-guided perturbation for unsupervised domain adaptation in segmentation.
\newblock In {\em Conference on Artificial Intelligence (AAAI)}, 2021.

\bibitem{wang2018deep}
Mei Wang and Weihong Deng.
\newblock Deep visual domain adaptation: A survey.
\newblock {\em Neurocomputing}, 312:135--153, 2018.

\bibitem{wang2021domain}
Qin Wang, Dengxin Dai, Lukas Hoyer, Luc Van~Gool, and Olga Fink.
\newblock Domain adaptive semantic segmentation with self-supervised depth estimation.
\newblock In {\em Proc. IEEE/CVF International Conference on Computer Vision}, pages 8515--8525, 2021.

\bibitem{wang2021exploring}
Wenguan Wang, Tianfei Zhou, Fisher Yu, Jifeng Dai, Ender Konukoglu, and Luc Van~Gool.
\newblock Exploring cross-image pixel contrast for semantic segmentation.
\newblock In {\em Proceedings of the IEEE/CVF International Conference on Computer Vision}, pages 7303--7313, 2021.

\bibitem{wang2021uncertainty}
Yuxi Wang, Junran Peng, and ZhaoXiang Zhang.
\newblock Uncertainty-aware pseudo label refinery for domain adaptive semantic segmentation.
\newblock In {\em Proc. IEEE/CVF International Conference on Computer Vision}, pages 9092--9101, 2021.

\bibitem{zhongDiff}
Zhonghao Wang, Mo You, Yunchao Wei, Rogerio Feris, Jinjun Xiong, Wen mei Hwu, Thomas~S. Huang, and Humphrey Shi.
\newblock Differential treatment for stuff and things: A simple unsupervised domain adaptation method for semantic segmentation.
\newblock In {\em IEEE Conference on Computer Vision and Pattern Recognition (CVPR)}, 2020.

\bibitem{wu2021one}
Xinyi Wu, Zhenyao Wu, Lili Ju, and Song Wang.
\newblock A one-stage domain adaptation network with image alignment for unsupervised nighttime semantic segmentation.
\newblock {\em IEEE Transactions on Pattern Analysis and Machine Intelligence}, 45(1):58--72, 2021.

\bibitem{xie2021segformer}
Enze Xie, Wenhai Wang, Zhiding Yu, Anima Anandkumar, Jose~M Alvarez, and Ping Luo.
\newblock Segformer: Simple and efficient design for semantic segmentation with transformers.
\newblock {\em Advances in Neural Information Processing Systems}, 34:12077--12090, 2021.

\bibitem{xu2021cdada}
Qi Xu, Yinan Ma, Jing Wu, Chengnian Long, and Xiaolin Huang.
\newblock Cdada: A curriculum domain adaptation for nighttime semantic segmentation.
\newblock In {\em IEEE Conference on Computer Vision and Pattern Recognition (CVPR)}, pages 2962--2971, 2021.

\bibitem{jinyuAdv}
Jinyu Yang, Chunyuan Li, Weizhi An, Hehuan Ma, Yuzhi Guo, Yu Rong, Peilin Zhao, and Junzhou Huang.
\newblock Exploring robustness of unsupervised domain adaptation in semantic segmentation.
\newblock In {\em IEEE International Conference on Computer Vision (ICCV)}, 2021.

\bibitem{yang2022divide}
Jianfei Yang, Xiangyu Peng, Kai Wang, Zheng Zhu, Jiashi Feng, Lihua Xie, and Yang You.
\newblock Divide to adapt: Mitigating confirmation bias for domain adaptation of black-box predictors.
\newblock {\em arXiv preprint arXiv:2205.14467}, 2022.

\bibitem{jihanAdv}
Jihan Yang, Ruijia Xu, Ruiyu Li, Xiaojuan Qi, Xiaoyong Shen, Guanbin Li, and Liang Lin.
\newblock An adversarial perturbation oriented domain adaptation approach for semantic segmentation.
\newblock In {\em Conference on Artificial Intelligence (AAAI)}, 2020.

\bibitem{yang2020fda}
Yanchao Yang and Stefano Soatto.
\newblock Fda: Fourier domain adaptation for semantic segmentation.
\newblock In {\em Proceedings of the IEEE/CVF Conference on Computer Vision and Pattern Recognition}, pages 4085--4095, 2020.

\bibitem{zhang2021prototypical}
Pan Zhang, Bo Zhang, Ting Zhang, Dong Chen, Yong Wang, and Fang Wen.
\newblock Prototypical pseudo label denoising and target structure learning for domain adaptive semantic segmentation.
\newblock In {\em Proc. IEEE/CVF conference on computer vision and pattern recognition}, pages 12414--12424, 2021.

\bibitem{yangCurriculum}
Yang Zhang, Philip David, Hassan Foroosh, and Boqing Gong.
\newblock A curriculum domain adaptation approach to the semantic segmentation of urban scenes.
\newblock {\em IEEE Transactions on Pattern Analysis and Machine Intelligence (PAMI)}, 42(8):1823--1841, 2020.

\bibitem{pspnet}
Hengshuang Zhao, Jianping Shi, Xiaojuan Qi, Xiaogang Wang, and Jiaya Jia.
\newblock Pyramid scene parsing network.
\newblock In {\em IEEE Conference on Computer Vision and Pattern Recognition (CVPR)}, 2017.

\bibitem{zhengRectify}
Zhedong Zheng and Yi Yang.
\newblock Rectifying pseudo label learning via uncertainty estimation for domain adaptive semantic segmentation.
\newblock {\em International Journal of Computer Vision (IJCV)}, 129(1):1106--1120, 2021.

\bibitem{zhou2022context}
Qianyu Zhou, Zhengyang Feng, Qiqi Gu, Jiangmiao Pang, Guangliang Cheng, Xuequan Lu, Jianping Shi, and Lizhuang Ma.
\newblock Context-aware mixup for domain adaptive semantic segmentation.
\newblock {\em IEEE Transactions on Circuits and Systems for Video Technology}, 2022.

\bibitem{zou2018unsupervised}
Yang Zou, Zhiding Yu, BVK Kumar, and Jinsong Wang.
\newblock Unsupervised domain adaptation for semantic segmentation via class-balanced self-training.
\newblock In {\em Proc. European Conference on Computer Vision (ECCV)}, pages 289--305, 2018.

\end{thebibliography}
}

\newpage

\clearpage
\setcounter{page}{1}
\maketitlesupplementary

%\maketitle

%%%%%%%%% BODY TEXT
\section{Supplementary Overview}
This is the supplementary material to support our manuscript "Unsupervised Domain Adaptation for Semantic Segmentation with Pseudo Label Self-Refinement". It contains additional quantitative and qualitative results related to our experiments that couldn't be included in the main article due to space constraints. In Sec.~\ref{sec:synth}, we provide quantitative results of SYNTHIA$\to$Cityscapes adaptation experiment comparing state-of-the-art methods. In Sec.~\ref{sec:stat}, we present ablation studies on Cityscapes$\to$Dark Zurich and SYNTHIA$\to$Cityscapes adaptation to analyze the impact of different components of our method. We also present experiments to show the effect of varying weights of refinement losses in this section. In Sec.~\ref{sec:qualit}, we showcase several qualitative examples of semantic segmentation, comparing our approach and baseline methods.

%statistics  related to our experiments that couldn't be included in the main article due to space constraints. 
%We present the  additional quantitative results in Sec.~\ref{sec:stat} and qualitative results in Sec.~\ref{sec:qualit}.

\renewcommand{\arraystretch}{1.4}
\begin{table*}[h]
\caption{Performance evaluation on \textbf{SYNTHIA$\to$Cityscapes} unsupervised domain adaptation. We report mIoU over 16 common categories between these datasets following prior works.}
\vspace{-1mm}
\resizebox{1\linewidth}{!}{
\begin{tabular}{c|c|cccccccccccccccc|c}
\toprule
\toprule
Method   &   & Road & S.Walk  & Build.  & Wall & Fence  & Pole & T.Light & Sign & Veget.  & Sky  & Person  & Rider   & Car  & Bus  & M.Bike  & Bike & \cellcolor{green!5}mIoU \\
\midrule
CBST ~\cite{zou2018unsupervised} & \parbox[t]{3mm}{\multirow{7}{*}{\rotatebox[origin=c]{90}{ResNet-Based}}} & 68.0 & 29.9 & 76.3 & 10.8 & 1.4 & 33.9 & 22.8 & 29.5 & 77.6 & 78.3 & 60.6 & 28.3 & 81.6 & 23.5 & 18.8 & 39.8 & \cellcolor{green!5}42.6 \\
CCM ~\cite{guanContent} & & 79.6 &36.4 &80.6 &13.3 &0.3 &25.5 &22.4 &14.9 &81.8 &77.4 &56.8 &25.9 &80.7 &45.3 &29.9 &52.0   &\cellcolor{green!5}45.2  \\
MetaCor ~\cite{guo2021metacorrection} & & 92.6 & 52.7 & 81.3 & 8.9 & 2.4 & 28.1 & 13.0 & 7.3 & 83.5 & 85.0 & 60.1 & 19.7 & 84.8 & 37.2 & 21.5 & 43.9 & \cellcolor{green!5}45.1 \\
DACS ~\cite{tranheden2021dacs} & & 80.6 & 25.1 & 81.9 & 21.5 & 2.9 & 37.2 & 22.7 & 24.0 & 83.7 & 90.8 & 67.6 & 38.3 & 82.9 & 38.9 & 28.5 & 47.6 & \cellcolor{green!5}48.4 \\
UAPLR ~\cite{wang2021uncertainty} & &    79.4 &34.6 &83.5 &19.3 &2.8 &35.3 &32.1 &26.9 &78.8 &79.6 &66.6 &30.3 &86.1 &36.6 &19.5 &56.9  & \cellcolor{green!5}48.0 \\
CorDA~\cite{wang2021domain} & & 93.3 & 61.6 & 85.3 & 19.6 & 5.1 & 37.8 & 36.6 & 42.8 & 84.9 & 90.4 & 69.7 & 41.8 & 85.6 & 38.4 & 32.6 & 53.9 & \cellcolor{green!5}55.0 \\
ProDA ~\cite{zhang2021prototypical} & & 87.8 & 45.7 & 84.6 & 37.1 & 0.6 & 44.0 & 54.6 & 37.0 & 88.1 & 84.4 & 74.2 & 24.3 & 88.2 & 51.1 & 40.5 & 45.6 & \cellcolor{green!5}55.5 \\
\textbf{DACS (w/ our PRN)}  &  & \textbf{88.1} & \textbf{47.1} & \textbf{84.8} & \textbf{37.5} & \textbf{0.9} & \textbf{45.0} & \textbf{55.4} & \textbf{38.6} & \textbf{88.2} & \textbf{85.2} & \textbf{75.2} & \textbf{25.5} & \textbf{88.4} & \textbf{51.9} & \textbf{41.3} & \textbf{46.4} & \cellcolor{green!5}\textbf{56.2} \\
\midrule
DAFormer~\cite{hoyer2022daformer} & \parbox[t]{3mm}{\multirow{5}{*}{\rotatebox[origin=c]{90}{SegFormer}}} & 84.5 & 40.7 & 88.4 & 41.5 & 6.5 & 50.0 & 55.0 & 54.6 & 86.0 & 89.8 & 73.2 & 48.2 & 87.2 & 53.2 & 53.9 & 61.7 & \cellcolor{green!5}60.9 \\
MIC-DAFormer~\cite{hoyer2023mic} &  &83.0 &40.9 &88.2 &37.6 &9.0 &52.4 &56.0 &56.5 &87.6 &93.4 &74.2 &51.4 &87.1 &59.6 &57.9 &61.2 &\cellcolor{green!5}62.2 \\
%Ours (SegFormer w/o CL \& FA) & 86.5 & 42.4 & 87.7 & 43.4 & 7.0 & 52.0 & 55.4 & 54.0 & 85.5 & 92.1 & 75.2 & 50.3 & 87.2 & 53.9 & 53.7 & 61.3 & \cellcolor{green!5}61.7 \\
\textbf{Ours}  &   & \textbf{86.6} & \textbf{44.7} & \textbf{91.7} & \textbf{44.4} & \textbf{9.3} & \textbf{53.0} & \textbf{55.9} & \textbf{57.2} & \textbf{88.3} & \textbf{89.2} & \textbf{75.1} & \textbf{49.8} & \textbf{91.2} & \textbf{56.9} & \textbf{55.9} & \textbf{63.8} & \cellcolor{green!5}\textbf{63.3} \\
\cline{1-1}\cline{3-19}
%Ours (SegFormer w/o CL \& FA) & 86.5 & 42.4 & 87.7 & 43.4 & 7.0 & 52.0 & 55.4 & 54.0 & 85.5 & 92.1 & 75.2 & 50.3 & 87.2 & 53.9 & 53.7 & 61.3 & \cellcolor{green!5}61.7 \\
%Ours (SegFormer w/o CL) & 87.9 & 43.6 & 90.7 & 44.8 & 9.0 & 49.2 & 58.0 & 57.1 & 88.4 & 92.7 & 75.2 & 50.2 & 89.0 & 55.5 & 56.1 & 62.7 & \cellcolor{green!5}63.1 \\
%Ours (SegFormer w/o FA) & 86.7 & 42.2 & 90.6 & 43.0 & 7.7 & 51.5 & 56.8 & 56.4 & 87.1 & 89.4 & 75.3 & 50.0 & 89.0 & 54.9 & 55.6 & 64.1 & \cellcolor{green!5}62.5   \\ 
DAFormer~(w/ HRDA)~\cite{hoyer2022hrda}  &   & 85.2	&47.7	&88.8	&49.5	&4.8	&57.2	&65.7	&60.9	&85.3		&92.9	&79.4	&52.8	&89.0		&64.7		&63.9	&64.9	 &\cellcolor{green!5}65.8  \\
\textbf{Ours~(w/ HRDA)}   &   & \textbf{87.8}	& \textbf{49.4}	&\textbf{88.1}	&\textbf{49.5}	&\textbf{5.3}	&\textbf{59.1}	&\textbf{65.6}	&\textbf{62.2}	&\textbf{85.6}		&\textbf{94.2}	&\textbf{79.1}	&\textbf{53.6}	&\textbf{87.1}		&\textbf{65.6}		&\textbf{65.8}	&\textbf{66.2}		
&\cellcolor{green!5}\textbf{66.5}  \\
\bottomrule
\bottomrule      
\end{tabular}
}
\label{tab:synthia}
\end{table*}
\renewcommand{\arraystretch}{1}

\vspace{0.2cm}
\section{SYNTHIA$\to$Cityscapes Results}
\label{sec:synth}
We compare our method with prior UDA methods on SYNTHIA$\to$Cityscapes adaptation in Table~\ref{tab:synthia}. From the last part of the table, it is evident that our method performs significantly better than SOTA methods (mIOU of $62.2$ with MIC-DAFormer and $60.9$ with DAFormer compared to $63.3$  with ours). Same as Cityscapes$\to$Dark Zurich and GTA$\to$Cityscapes results in the main paper, our method consistently achieves higher IoU across most classes. The ResNet-based baseline DACS (w/ our PRN) was trained by combining our PRN module with the prior method DACS. We train this baseline to compare with prior pseudo-label selection or refinement-based UDA methods reported in the first part of Table~\ref{tab:synthia} (e.g., CCM, MetaCor, UAPLR, ProDA). We see our PRN module leads to significant improvement over other pseudo-label selection or refinement-based UDA methods. From the last part of Table~\ref{tab:synthia}, we see incorporating HRDA training leads to further improvement in performance, and Ours with HRDA performs better than state-of-the-art DAFormer with HRDA.

\vspace{0.2cm}
\section{Ablation Studies}
\label{sec:stat}
We have presented the ablation study of our proposed method on GTA$\to$Cityscapes in Table 3 of the main paper. Here, we present an ablation study on Cityscapes$\to$Dark-Zurich in Table~\ref{tab:ablation_dz} to analyze different components of our method, i.e., Self-Training (ST), Pseudo
Label Refinement (PL-R), Noise Mask (NM), Contrastive Learning without or with using the output of PRN (CL w/o
R, CL w/ R) and Fourier Adaptation (FA).  We again observe that the proposed method leads to a large improvement over the self-training baseline reported in the second row ($58.4$ in row-2.8 vs. $51.2$ in row-2.2). We also observe that our proposed PRN module (with both pseudo-label refinement and noise-mask prediction) leads to significant improvement over the self-training baseline (row-2.4 vs. row-2.2). The impact of noise-mask prediction in PRN shows improvement compared to without it (row-2,4 vs. row-2.2). It is also evident that our pseudo-label refinement is crucial to achieving a performance boost with the contrastive learning module comparing row-2.5 and row-2.6 with row-2.4. We see the use of the PRN module output is crucial for contrastive learning to achieve a performance boost. Comparing row-2.7 with row-2.4, we see performance improvement by applying the FA module. Finally, row-2.8 shows the performance when all the components of our framework are used.

\vspace{0.1cm}
We also perform an ablation study on SYNTHIA$\to$Cityscapes in Table~\ref{tab:ablation_synthia}. We observe a similar trend to Cityscapes$\to$Dark-Zurich and GTA5$\to$Cityscapes ablation studies that different components of the proposed UDA framework with pseudo-label refinement module consistently help improve performance significantly.

\renewcommand{\arraystretch}{1.35}
\begin{table}[t]
\caption{Ablation study with different components of our proposed method on \textbf{Cityscapes$\to$Dark-Zurich}.}
\resizebox{1\linewidth}{!}{
\begin{tabular}{c|cccccc|c}
\toprule
\toprule
\# & ST & PL-R & NM & CL w/o R & CL w/ R & FA & mIoU \\
\midrule
2.1 & x    & x    & x     & x    & x   & x  & 37.5 \\
2.2 & \checkmark    & x    & x     & x    & x   & x  & 51.2 \\
2.3 & \checkmark    &  \checkmark    & x     & x    & x   & x  & 54.9 \\
2.4 & \checkmark    &  \checkmark    &  \checkmark     & x    & x   & x  & 55.8 \\
2.5 & \checkmark    &  \checkmark    &  \checkmark     &  \checkmark    & x   & x  & 55.3 \\
2.6 & \checkmark    &  \checkmark    &  \checkmark     & x    &  \checkmark   & x  & 56.5 \\
2.7 & \checkmark    &  \checkmark    &  \checkmark     & x    & x   &  \checkmark  & 58.0   \\
\midrule
\rowcolor{aliceblue} 2.8 & \checkmark    &  \checkmark    &  \checkmark     & x    &  \checkmark   &  \checkmark  & 58.4 \\
\bottomrule
\bottomrule
\end{tabular}
}
\label{tab:ablation_dz}
\end{table}

\renewcommand{\arraystretch}{1}

\iffalse
\begin{table}[h]
\caption{Ablation study with different components of our proposed method on \textbf{GTA5$\to$Cityscapes}.}
\vspace{1mm}
\resizebox{1\linewidth}{!}{
\begin{tabular}{c|cccccc|c}
\toprule
\# & ST & PL-R & NM & CL w/o R & CL w/ R & FA & mIoU \\
\midrule
1.1 & x    & x    & x     & x    & x   & x  & 45.6 \\
1.2 & \checkmark    & x    & x     & x    & x   & x  & 67.3 \\
1.3 & \checkmark    &  \checkmark    & x     & x    & x   & x  & 69.7 \\
1.4 & \checkmark    &  \checkmark    &  \checkmark     & x    & x   & x  & 70.1 \\
1.5 & \checkmark    &  \checkmark    &  \checkmark     &  \checkmark    & x   & x  & 70.1 \\
1.6 & \checkmark    &  \checkmark    &  \checkmark     & x    &  \checkmark   & x  & 70.5 \\
1.7 & \checkmark    &  \checkmark    &  \checkmark     & x    & x   &  \checkmark  & 70.8   \\
\midrule
\rowcolor{aliceblue} 1.8 &  \checkmark    &  \checkmark    &  \checkmark     & x    &  \checkmark   &  \checkmark  & 71.3 \\
\bottomrule
\end{tabular}
}
\label{tab:ablation_gta}
\end{table}
\fi

\begin{figure}[t]
    \centering
    \includegraphics[width=0.95\linewidth]{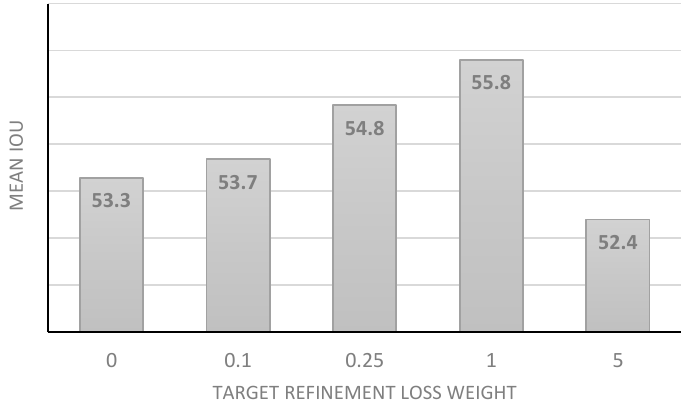}
    \caption{Results on varying weight for target refinement loss (i.e., $\mathcal{L}^{RT}_{ce}$ + $\mathcal{L}^{RT}_{bce}$), while keeping the weight (i.e., $\lambda_2$) of source refinement loss fixed in Cityscapes$\to$Dark-Zurich. For this experiment, we use our proposed model without the CL \& FA components.}
    \label{fig:target_loss_stats}
\end{figure}

%\newpage
\vspace{0.1cm}
In Fig.~\ref{fig:target_loss_stats}, we present results on Cityscapes$\to$Dark-Zurich by varying weight for target refinement loss (i.e., $\mathcal{L}^{RT}_{ce}$ + $\mathcal{L}^{RT}_{bce}$), while keeping the weight (i.e., $\lambda_2$) of source refinement loss (i.e., $\mathcal{L}^{RS}_{ce}$ + $\mathcal{L}^{RS}_{bce}$) fixed. For this experiment, we use our proposed model without the additional CL and FA modules (i.e., row-2.4 of Table.~\ref{tab:ablation_dz}).  As reported in row-2.2 of Table 2, the self-training baseline achieves mIoU of $51.2$. We observe mIoU improvement compared to the self-training baseline in all the cases.  When the target pseudo-label refinement loss is not used (i.e., weight is set to $0$), the performance drops to mIoU of $53.3$ ($-2.5\%$ compared to the case of loss weight set to $1$). It shows that the source refinement loss is effective in improving pseudo-label quality and overall performance ($53.3$ vs. the self-training baseline result of $51.2$). However, the target refinement loss helps to further improve the performance. The best performance is achieved with the target refinement loss weight set to $1$.

\renewcommand{\arraystretch}{1.35}
\begin{table}[t]
\caption{Ablation study with different components of our proposed method on \textbf{SYNTHIA$\to$Cityscapes}.}
\resizebox{1\linewidth}{!}{
\begin{tabular}{c|cccccc|c}
\toprule
\toprule
\# & ST & PL-R & NM & CL w/o R & CL w/ R & FA & mIoU \\
\midrule
3.1 & x    & x    & x     & x    & x   & x  & 46.5 \\
3.2 & \checkmark    & x    & x     & x    & x   & x  & 60.9 \\
3.3 & \checkmark    &  \checkmark    & x     & x    & x   & x  & 61.7 \\
3.4 & \checkmark    &  \checkmark    &  \checkmark     & x    & x   & x  & 62.1 \\
3.5 & \checkmark    &  \checkmark    &  \checkmark     &  \checkmark    & x   & x  & 62.2 \\
3.6 & \checkmark    &  \checkmark    &  \checkmark     & x    &  \checkmark   & x  & 62.5 \\
3.7 & \checkmark    &  \checkmark    &  \checkmark     & x    & x   &  \checkmark  & 63.1   \\
\midrule
\rowcolor{aliceblue}  3.8 & \checkmark    &  \checkmark    &  \checkmark     & x    &  \checkmark   &  \checkmark  & 63.3 \\
\bottomrule
\bottomrule
\end{tabular}
}
\label{tab:ablation_synthia}
\end{table}
\renewcommand{\arraystretch}{1}

\vspace{0.2cm}
\section{Qualitative Results}
\label{sec:qualit}
\begin{figure*}[t]
    \centering
    \includegraphics[width=0.96\linewidth]{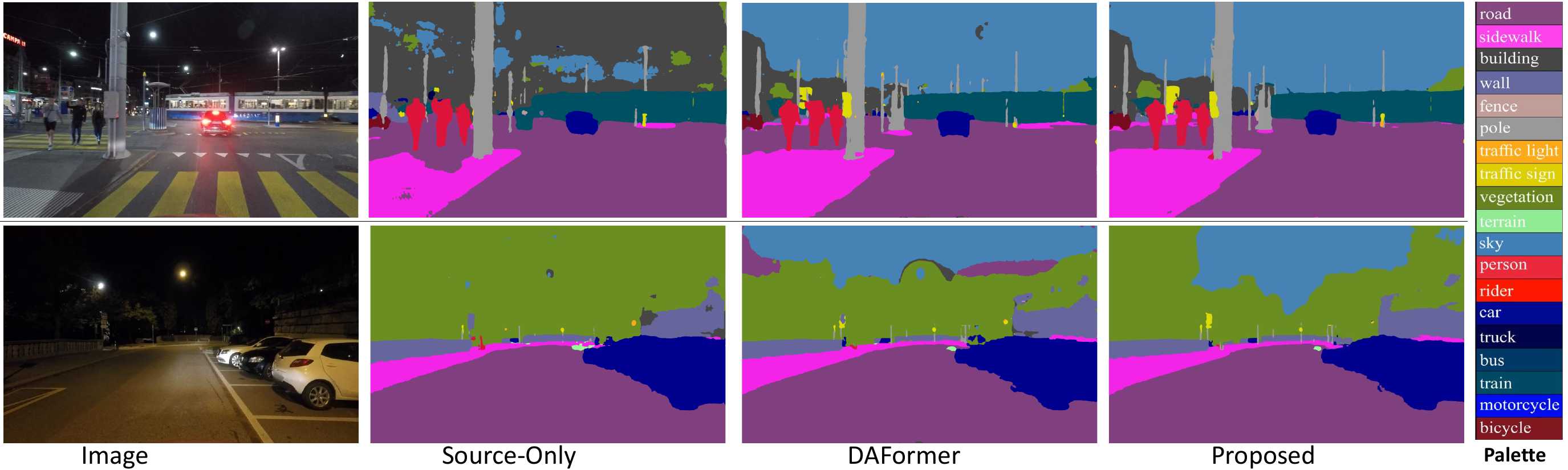}
    \caption{Qualitative Examples of Cityscapes$\to$Dark-Zurich on Dark Zurich test set comparing the Source-Only baseline and DaFormer.}
    \label{fig:qualit_dz}
\end{figure*}

\begin{figure*}[t]
    \centering
    \vspace{2mm}
    \includegraphics[width=0.96\linewidth]{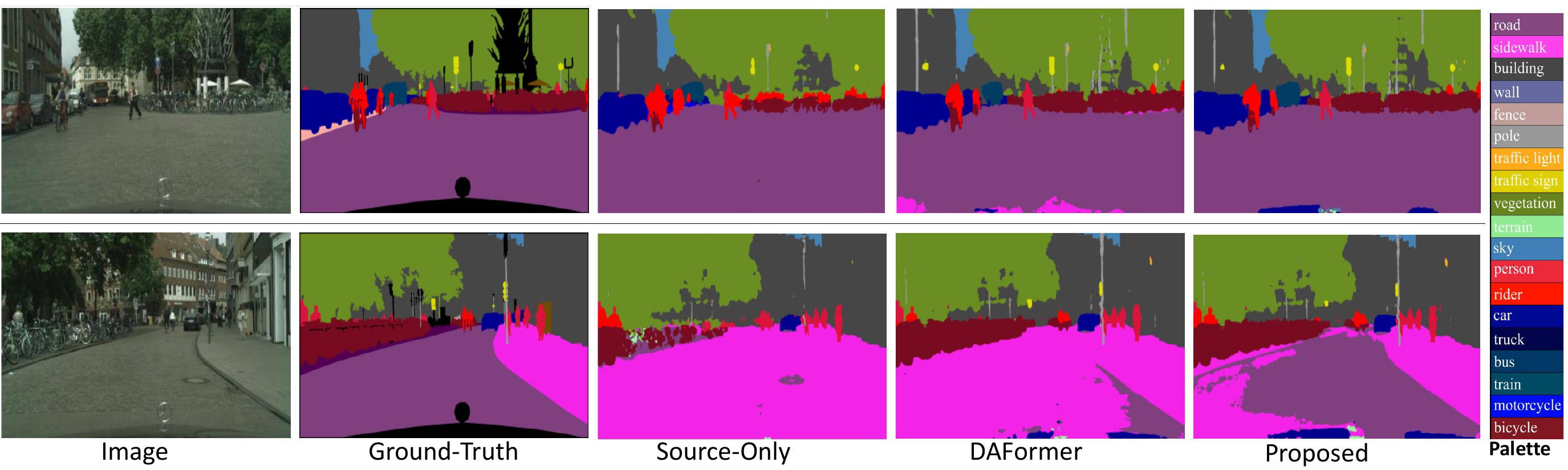}
    \caption{Qualitative Evaluation on GTA5$\to$Cityscapes on Cityscapes val. set comparing GT, Source-Only baseline and DaFormer.}
    \label{fig:qualit_cs}
\end{figure*}

In this section, we present the qualitative comparison of our approach with the state-of-the-art method DAFormer. The Source-Only baseline results (with no domain adaptation) are also shown for reference. Fig.~\ref{fig:qualit_dz} shows qualitative examples of our method in adapting the model trained on Cityscapes to Dark-Zurich. Similar to the qualitative examples presented in the main paper, we again see that our approach leads to a significant improvement in several classes which can be hard to classify due to changes in domains. We couldn't show the ground truth label in Fig.~\ref{fig:qualit_dz} as we do not have direct access to it for the test set of Dark-Zurich. Fig.~\ref{fig:qualit_cs} shows the qualitative results for adaptation from GTA5 to Cityscapes. These results also include the ground truth semantic labels for reference. We again qualitatively observe that our proposed method consistently performs better than the compared methods.

%{\small
%\bibliographystyle{ieee_fullname}
%\bibliography{egbib}
%}

\end{document}